%% file: root.tex
\documentclass[letterpaper, 10 pt, conference]{ieeeconf}  

\IEEEoverridecommandlockouts                              

\usepackage[T1]{fontenc}
\usepackage[greek, english]{babel}
\usepackage{color}%
\usepackage[pdftex]{graphicx}
\usepackage{random}
\usepackage{amsfonts} 
\usepackage{amssymb} 
\usepackage{amsmath} 
\usepackage{amsbsy} 
\usepackage{fixmath}
\usepackage{import}

\title{\LARGE \bf
Probability Estimation for Predicted-Occupancy Grids
in Vehicle Safety Applications Based on Machine Learning
\thanks{©~2016 IEEE. Personal use of this material is permitted. Permission from IEEE must be obtained for all other uses, in any current or future media, including reprinting/republishing this material for advertising or promotional purposes, creating new collective works, for resale or redistribution to servers or lists, or reuse of any copyrighted component of this work in other works. 
This is the author's accepted manuscript version of the paper: “Probability estimation for Predicted-Occupancy Grids in vehicle safety applications based on machine learning” by P.~Nadarajan and M.~Botsch, published in the 2016 IEEE Intelligent Vehicles Symposium (IV), Gothenburg, Sweden, pp.~1285--1292, 2016. DOI: 10.1109/IVS.2016.7535556.}}

\author{Parthasarathy Nadarajan$^{1}$ and Michael Botsch$^{1}$
\thanks{$^{1}$Technische Hochschule Ingolstadt, Esplanade 10
        Ingolstadt, Germany}
}%

\begin{document}

\maketitle
\thispagestyle{empty}
\pagestyle{empty}

\begin{abstract}

This paper presents a method to predict the evolution of a complex
traffic scenario with multiple objects. The current state of the
scenario is assumed to be known from sensors and the prediction is
taking into account various hypotheses about the behavior of
traffic participants. This way, the uncertainties regarding the
behavior of traffic participants can be modelled in detail. In the
first part of this paper a model-based approach is presented to
compute \emph{Predicted-Occupancy Grids} (POG), which are
introduced as a grid-based probabilistic representation of the
future scenario hypotheses. However, due to the large number of
possible trajectories for each traffic participant, the
model-based approach comes with a very high computational load.
Thus, a machine-learning approach is adopted for the computation
of POGs. This work uses a novel grid-based representation of the
current state of the traffic scenario and performs the mapping to
POGs. This representation consists of augmented cells in an
occupancy grid. The adopted machine-learning approach is based on
the Random Forest algorithm. Simulations of traffic scenarios are
performed to compare the machine-learning with the model-based
approach. The results are promising and could enable the real-time
computation of POGs for vehicle safety applications. With this
detailed modelling of uncertainties, crucial components in vehicle
safety systems like criticality estimation and trajectory planning
can be improved.

\end{abstract}

\section{INTRODUCTION}

In recent years the field of Active Vehicle Safety gained
significant importance due to the availability of sensors like
radar, cameras, laserscanners, etc. Using the information of such
exteroceptive sensors, vehicles are capable of estimating the
criticality of traffic scenarios and to avoid or mitigate
collisions. For example, the Autonomous Emergency
Braking~\cite{Kaempchen09} is an important Active Vehicle Safety
system that is already available on the market. One of the major
challenges for Active Vehicle Safety systems is the prediction on
how a traffic scenario can evolve in a given time horizon.
Predictions over a short interval of time, which for most safety
applications is less than approx. $1$\,s, can be estimated based
on the current state of the traffic objects~\cite{Herrmann15}.
These states contain physical quantities like velocities, yaw
rates, accelerations etc. However, predictions over longer
horizons depend largely on the interaction and the motivation of
the traffic participants in a particular traffic scene.

In recent years several contributions were made in anticipating
the behavior of traffic participants by incorporating the
intention of the driver and the interaction between the
participants. For longer periods of prediction, the methods can be
categorized into pattern recognition in motion pattern
databases~\cite{Hermes09, Vasquez04} and methods which fuse
dynamic motion models with behavior and environment
description~\cite{Broadhurst05}. There are also a
number of approaches which predict the evolution of a traffic
scene on a more abstract level. The models consist of
probabilistic state machines~\cite{Hulnhagen10}, static Bayesian
Networks~\cite{Kasper12}, dynamic Bayesian
Networks~\cite{Dagli03}, Hidden Markov Models~\cite{Meyer09} or
fuzzy-theory~\cite{Hulnhagen10}. In~\cite{Schreier14}, long-term
trajectory prediction is based on a combination of high-level Bayesian maneuver detection
with immanent uncertainty
of maneuver execution by drivers. In~\cite{Gindele10} and~\cite{Lefevre12}, a dynamic Bayesian
Network is used to model the interaction between an arbitrary
number of traffic participants.

In order to make accurate predictions, the representation of the
environment plays an important role. The objects in the
environment vary a lot with respect to their geometries and
description of these objects can be quite complex. One of the
established methods to overcome this problem is the occupancy grid
framework~\cite{Thrun05}. The occupancy grid divides the
environment into a grid of cells and the occupancy of each of the
cells being occupied or empty is estimated. Occupancy grids find
their application in sensor data fusion~\cite{Stepan05}, path
planning~\cite{Himmelsbach08}, simultaneous localization and
mapping (SLAM)~\cite{Siciliano08} and target
tracking~\cite{Chen06}. Work also has been done for the
representation of dynamic situations with occupancy grids. The
basic Bayesian occupancy filter combines the occupancy grid
representation of static environments with probabilistic velocity
objects to build a dynamic map of the environment.
In~\cite{Chen06}, a different formulation with a 2-D occupancy,
where each cell has an associated distribution over its possible
velocities is adpoted.

In this paper, a novel method for the representation and
prediction of traffic scenes using occupancy grids is formulated.
The main idea is to augment the cells with information about the
infrastructure and traffic participants, e.\,g., acceleration,
velocity and yaw angle, and then to use this representation for
predictions. The result is called \emph{Predicted-Occupancy Grid}
(POG) and includes the uncertainty about the future motion of
traffic objects. Due to the high computational complexity
necessary to model and evaluate a large number of motion
hypotheses for each object, the approach introduced in this work
to enable real-time computation is based on machine learning. The
evolution of a traffic scenario is estimated as a Predicted-Occupancy
grid using the \emph{Random Forest} (RF)
algorithm~\cite{breiman01random}.

The outline of the paper is as follows.
Section~\ref{sec:Predicted-Occupancy} introduces the concept of
POGs. Section~\ref{sec:Machine Learning} deals with the machine
learning technique used for the probability estimation for POGs.
The evaluation of the machine-learning approach with simulation
results is presented in Section~\ref{sec:Evaluation}. The
applications of the current approach towards the field of Vehicle
Safety is discussed in Section~\ref{sec:Applications}.

Throughout this work, vectors and matrices are denoted by lower
and upper case bold letters, and random variables are written
using sans serif fonts. A lower-case bold letter represents a
column vector.

\section{Predicted-Occupancy Grids}
\label{sec:Predicted-Occupancy} In~\ref{subsec:DynamicModels}
and~\ref{subsec:PredictedHypothesis}, the theoretical background
for the development of a model-based approach for the computation
of POG is presented. The models will also be used for the
generation of a training data set for the machine-learning
approach.

\subsection{Dynamic Models of Traffic Objects}
\label{subsec:DynamicModels} To construct POGs, a fundamental task
is the motion-prediction of objects in traffic scenarios. This
prediction requires dynamic models which are used to compute
future trajectories. In~\cite{Schubert08} a survey on dynamic
vehicle models and their performance for tracking tasks are
presented. For tracking tasks, simple linear models like the
Constant Velocity (CV) or Constant Acceleration (CA) models can be
used. But for a better performance, curvilinear models like Constant
Turn Rate and Velocity (CTRV), Constant Turn Rate and Acceleration
(CTRA), Constant Steering Angle and Velocity (CSAV), or Constant
Curvature and Acceleration (CCA) are used.

Since the focus in this paper is to predict the planar motion of
objects in traffic scenarios with high accuracy, more
sophisticated models than the ones presented in~\cite{Schubert08}
are used. The CV, CA, CTRV, CTRA, CSAV, and CCA models are
simplifications of the single-track vehicle model, that is often
called bicycle model. The single-track model is a
simplification of the two-track model which is presented in the
following.

\subsubsection{Two-Track Model}%
\begin{figure}[htb!]
    \begin{scriptsize}
    \vspace{-0.2cm}
    \centering
    \def\svgwidth{0.4\textwidth}
    \centerline{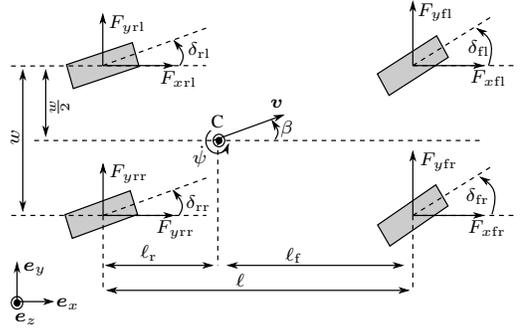}
    \end{scriptsize}
    \vspace{-0.4cm}
\def\figurename{Figure}
\caption{Two-track model}%
\vspace{-0.2cm}%
\label{FigTwoTrackModel}
\end{figure}

The planar motion of $4$-wheel cars can be described accurately
using a two-track model. It is a nonlinear planar vehicle dynamic
model which describes the vehicle as a rigid body and models the
forces that act on the $4$ tires. Fig.~\ref{FigTwoTrackModel}
shows the planar view of a vehicle and presents the quantities
that are needed to derive the two-track model starting from the
balance of forces and moments.
The vehicle coordinate system has the unit vectors $\boldsymbol{e}_x$,
$\boldsymbol{e}_y$, and $\boldsymbol{e}_z$. The center of gravity is denoted with
$C$, the track width with $w$, the wheel-base with $\ell$, the
slip angle with $\beta$, the velocity vector with $\boldsymbol{v}$, the
yaw-rate with $\dot{\psi}$, and the steering angles at each wheel
with $\delta_{i}$, where the subscript $i\in\{\text{fl, fr, rl,
rr}\}$ indicates whether it is the front left (fl), front right
(fr), rear left (rl) or rear right (rr) wheel. The same subscript
notation is used for the forces $F_{xi}$ and $F_{yi}$ acting on
each wheel. Denoting the velocity magnitude with $v$ the velocity
vector $\boldsymbol{v}$ can be written as
\begin{align}
\boldsymbol{v}=v_x\boldsymbol{e}_x+v_y\boldsymbol{e}_x=v\cos(\beta)\boldsymbol{e}_x+v\sin(\beta)\boldsymbol{e}_x.
\end{align}
The acceleration $\boldsymbol{a}=\dot{\boldsymbol{v}}$ results by taking into
account that the vehicle coordinate frame is a rotating frame with
angular rate $\dot{\psi}$ and this leads to
\begin{align}
\boldsymbol{a}&=\left(\dot{v}\cos(\beta) - v \left(\dot{\beta} + \dot{\psi}\right)\sin(\beta)\right)\boldsymbol{e}_{x} \nonumber \\
            &\quad + \left(\dot{v}\sin(\beta) + v \left(\dot{\beta} +  \dot{\psi}\right)\cos(\beta) \right)\boldsymbol{e}_{y}
\end{align}
Thus, the balance of forces for the x-direction is
\begin{align}
    F_{x\mathrm{fl}}\!+\!F_{x\mathrm{fr}}\!+\!F_{x\mathrm{rl}}
    \!+\!F_{x\mathrm{rr}}&\!=\!m \!\left(\!\dot{v}\cos(\beta) \!-\! v
    \!\left(\!\dot{\beta} \!+\! \dot{\psi}\!\right)\!\sin(\beta)\!\right)\!, \label{EqSumFxTwoTrack}
\end{align}
and for the y-direction
\begin{align}
    F_{y\mathrm{fl}}\!+\!F_{y\mathrm{fr}}\!+\!F_{y\mathrm{rl}}\!+\!F_{y\mathrm{rr}}
    &\!\!=m \!\left(\!\dot{v}\sin(\beta) \!+\! v \!\left(\!\dot{\beta} \!+\!
    \dot{\psi}\!\right)\!\cos(\beta) \!\right)\!. \label{EqSumFyTwoTrack}
\end{align}
Multiplying Eq.~(\ref{EqSumFxTwoTrack}) with $\cos(\beta)$ and
Eq.~(\ref{EqSumFyTwoTrack}) with $\sin(\beta)$ and adding the
results leads to the first differential equation of the two-track
model
\begin{align}
\dot{v}=\frac{1}{m}\left(\cos(\beta)\sum\limits_{i}
F_{xi}+\sin(\beta)\sum\limits_{i} F_{yi}  \right).
\label{EqDotVtwoTrack}
\end{align}
The second differential equation of the two-track model results
from introducing Eq.~(\ref{EqDotVtwoTrack}) in
Eq.~(\ref{EqSumFyTwoTrack})
\begin{align}
    \dot{\beta}=\frac{1}{mv}\left(\cos(\beta)\sum\limits_{i} F_{yi} -
    \sin(\beta) \sum\limits_{i} F_{xi}\right) - \dot{\psi}.
    \label{EqDotSliptwoTrack}
\end{align}
The third and last differential equation of the two-track model
results from the balance of moments
\begin{align}
    \sum\limits_{i} \boldsymbol{r}_i\times \boldsymbol{F}_i=I_z \dot{\boldsymbol{\omega}},
    \label{EqSumMomentsBodyCoord}
\end{align}
where $\boldsymbol{r}_i$ is the lever arm for the force $\boldsymbol{F}_i$ with
respect to $C$, $I_z$ is the yaw moment of inertia and
$\dot{\boldsymbol{\omega}}=[0,0,\ddot{\psi}]^{\text{T}}$. The differential
equation is
\begin{align}
    \ddot{\psi}&=\frac{1}{I_z}\big(\ell_{\mathrm{f}}(F_{y\mathrm{fl}}+F_{y\mathrm{fr}})+ \frac{w}{2}(F_{x\mathrm{fr}}-F_{x\mathrm{fl}}) \nonumber\\
    &\quad - \ell_{\mathrm{r}}((F_{y\mathrm{rl}}+F_{y\mathrm{rr}}) +
    \frac{w}{2}(F_{x\mathrm{rr}}-F_{x\mathrm{rl}}) \big).
    \label{EqDotYawtwoTrack}
\end{align}
In order to generate motion predictions for a $4$-wheel car,
suitable input quantities are needed whose effects on the forces
$F_{xi}$ and $F_{yi}$ can easily be modelled. Using such input
quantities that model the forces on each wheel, numerical
integration of Eqs.~(\ref{EqDotVtwoTrack}),
(\ref{EqDotSliptwoTrack}), and (\ref{EqDotYawtwoTrack}) leads to
the quantities $\dot{\psi}$, $\beta$, $v$ and by further
integration and coordinate transformation to the position $X$ and
$Y$ of the center of gravity $C$ and to the yaw-angle $\psi$ in
the global coordinate frame. This way the planar location and
orientation of the vehicle can be computed as a result of the
chosen input quantities. Simple and intuitive input quantities
that can be used to model the forces on each wheel are the
steering-wheel angle and the position of the gas pedal and of the
braking pedal. To use these inputs a decomposition of the forces
$F_{xi}$ and $F_{yi}$ in the local coordinate frame of the
$i\text{-th}$ wheel is introduced, since the dependence of these
forces on the longitudinal and side slip can be described
mathematically. This nonlinear dependence is modelled by the
``magic tire formula'' that is presented in~\cite{Bakker87}. The
longitudinal slip can be modelled to be a function of the gas
pedal and of the brake pedal positions. The side slip can be
modelled to be a function of the steering-wheel
angle~\cite{Baffet08}. Thus, the planar location and orientation
of the vehicle can be predicted based on hypotheses about the
steering-wheel angle and brake and gas pedal positions. Such a
driver model for predictions in traffic scenarios is implemented
for this work and each hypothesis of the driver inputs leads to a
predicted trajectory.

\subsubsection{Single-Track Model}%
The single-track model is a simplification of the the two-track
model which results from the assumptions that the vehicle width
$w=0$ and the slip-angle $\beta$ is small~\cite{Jazar09}. This
model can be used to describe the dynamics of cars for low values
of the lateral acceleration. In this work it will be used to make
predictions for bicycles in traffic scenarios. The input
quantities that are used in this work for the generation of
trajectories using the bicycle-model are the longitudinal
acceleration and the steering angle.

By variations of input quantities mentioned above, it is
possible to generate multiple hypotheses starting from an initial
state of a traffic participant. The next section deals with the
model used in this work to generate multiple hypotheses.

\subsection{Prediction and Generation of Hypothesis}
\label{subsec:PredictedHypothesis} The aim for introducing
multiple hypotheses is to generate a detailed model of the
uncertainty related to the future behavior of traffic
participants.

\begin{figure}[htb!]
    \begin{scriptsize}
    \centering
    \def\svgwidth{0.4\textwidth}
    \centerline{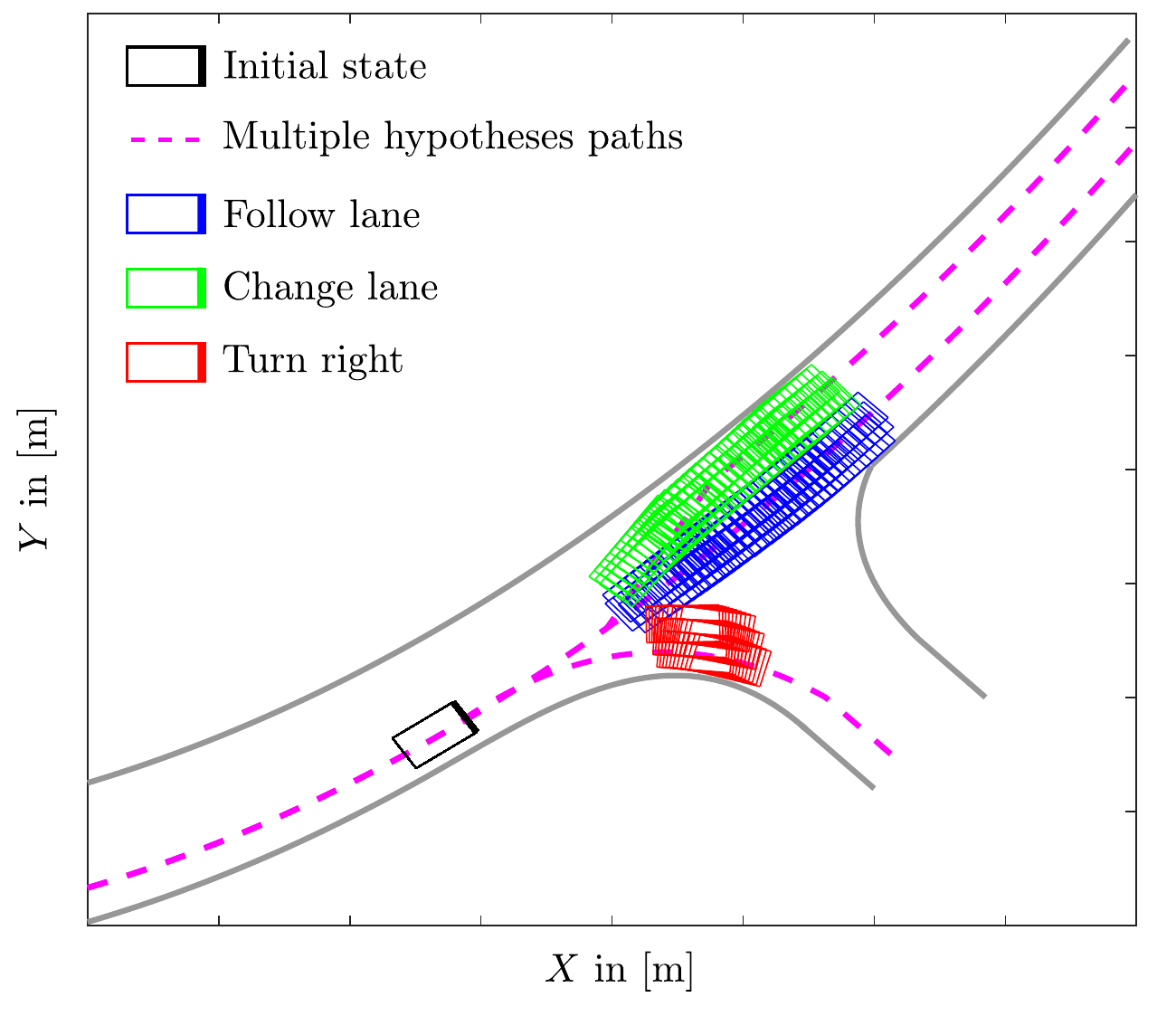}
    \end{scriptsize}
    \vspace{-0.2cm}
\def\figurename{Figure}
\caption{Multiple Hypothesis}%
\vspace{-0.6cm}%
\label{FigMultipleHypothesis}
\end{figure}

The prediction hypotheses are categorized into two viz. the
\emph{main hypothesis} (MH) which describes a possible main
maneuver, e.\,g., follow lane, drive straight, turn left, turn
right, change lane, etc. and the \emph{sub hypothesis} (SH) which
includes the dynamic uncertainties during the execution of a
certain main hypothesis, e.\,g., lateral and longitudinal
deviation. A \emph{main hypothesis} does not consider any abnormal
driver behavior, e.\,g., moving out of the road.
Fig.~\ref{FigMultipleHypothesis} shows a scenario with $3$ main
hypotheses such as follow lane (blue), change lane (green) and
turn right (red) and a large number of \emph{sub hypotheses} per
\emph{main hypothesis}.
\subsubsection{Main Hypothesis}
The main hypotheses for the vehicle $\text{V}_\ell$ are modelled
as a discrete random variable $\random{h}^{(MH)}_{\text{V}_\ell}$
with the outcomes $\{$follow lane, drive straight, change lane,
turn right, turn left, $\ldots\}$.
In this work the assignment of the probability for
$\random{h}^{(MH)}_{\text{V}_\ell}$ is rule-based and relies on
expert knowledge. This assignment takes into account the current
state vectors of all $L$ traffic participants and
information about the road infrastructure and traffic rules.
Hereby, the state vector of the $\ell\text{-th}$ vehicle
\begin{align}
\mathbf{x}_{\textrm{V}_{\ell}} =
[X_{\ell},Y_{\ell},v_{\ell},\psi_{\ell},a_{x,\ell},a_{y,\ell}]^T
\end{align}
comprises the $X_{\ell}$ and $Y_{\ell}$ positions of the center of
gravity $C$ in the global coordinate frame, velocity $v_{\ell}$, orientation $\psi_{\ell}$, longitudinal
$a_{x,\ell}$ and lateral $a_{y,\ell}$ acceleration of the vehicle.
For example in the scenario considered in
Fig.~\ref{FigMultipleHypothesis}, it is more likely that the
vehicle would make a right turn when it decelerates or has a lower
velocity and it is more likely to follow the lane when the vehicle
is neither accelerating nor decelerating.
\\At every time instance of the
prediction horizon the sum of the $M$ \emph{main hypothesis} that
are modelled for the vehicle $V_\ell$, should be equal to $1$
\begin{equation}
\sum_{m=1}^{M}
\text{P}\left(\random{h}^{(MH)}_{\text{V}_\ell}=h^{(MH)}_{\text{V}_\ell,m}\right)=1,
\end{equation}
where $h^{(MH)}_{\text{V}_\ell,m}$ is the $m$-th possible outcome
of $\random{h}^{(MH)}_{\text{V}_\ell}$.

\subsubsection{Sub Hypothesis}
Each of the \emph{main hypotheses} will branch into multiple
\emph{sub hypotheses}. At the prediction time instance $t_\text{pred}$, \emph{sub hypotheses} corresponding to the main hypothesis
$h^{(MH)}_{\text{V}_\ell,m}$ are modelled as a multivariate random
variable $\Brandom{h}^{(SH/m)}_{\text{V}_\ell,t_\text{pred}}$ to
include the uncertainties of the future driver behavior concerning
the longitudinal and lateral dynamics. The random variable
$\Brandom{h}^{(SH/m)}_{\text{V}_\ell,t_\text{pred}}$ consists of
two elements
\begin{align}
\Brandom{h}^{(SH/m)}_{\text{V}_\ell,t_\text{pred}}=\left[
\begin{array}{c}
\random{d}^{(SH/m)}_{\text{V}_\ell,t_\text{pred},\text{lon}}\\
\random{d}^{(SH/m)}_{\text{V}_\ell,t_\text{pred},\text{lat}}
\end{array}
\right].
\end{align}
The first element of
$\Brandom{h}^{(SH/m)}_{\text{V}_\ell,t_\text{pred}}$ is the
deviation in longitudinal direction of the vehicle $V_\ell$ at
time instance $t_\text{pred}$ from its location according to the
corresponding main hypothesis $m$ at the same prediction instance
$t_\text{pred}$. The second element of
$\Brandom{h}^{(SH/m)}_{\text{V}_\ell,t_\text{pred}}$ is defined
equivalently for the lateral deviation.

To take into account that large deviations are unlikely, the
conditional probability
$\text{p}\left(\random{d}^{(SH/m)}_{\text{V}_\ell,t_\text{pred},\text{lon}}
\vert h^{(MH)}_{\text{V}_\ell,m}\right)$ is modelled as a
triangular \emph{probability density function} (pdf). The bounds
of the triangular pdf for the longitudinal deviation are functions
of the prediction time, the velocity and the acceleration of the
vehicle. The lower bound of the longitudinal deviation will be
attained at the maximum deceleration $a_\text{decel,max}$ of the
traffic object. Similarly, the upper bound will be attained with
the maximum acceleration limit of the traffic object
$a_\text{accel,max}$. Equivalently, the conditional probability
$\text{p}\left(\random{d}^{(SH/m)}_{\text{V}_\ell,t_\text{pred},\text{lat}}\vert
h^{(MH)}_{\text{V}_\ell,m}\right)$ is modelled as a triangular
distribution considering that large deviations are less likely.
The bounds of this pdf are functions of the prediction time, the
road limits, the velocity and the lateral acceleration. The
maximum magnitude for the lateral acceleration is denoted
with $a_\text{lat,max}$.

A sub-hypothesis, i.\,e., a driving maneuver branching from the
$m\text{-th}$ main hypothesis, is a realization of the joint
conditional pdf
$\text{p}\left(\random{d}^{(SH/m)}_{\text{V}_\ell,t_\text{pred},\text{lon}},
\random{d}^{(SH/m)}_{\text{V}_\ell,t_\text{pred},\text{lat}} \vert
h^{(MH)}_{\text{V}_\ell,m}\right)$ which can be decomposed
assuming statistical independence between lateral and longitudinal
deviations as
\begin{align}
\text{p}&\left(\random{d}^{(SH/m)}_{\text{V}_\ell,t_\text{pred},\text{lon}},
\random{d}^{(SH/m)}_{\text{V}_\ell,t_\text{pred},\text{lat}} \vert
h^{(MH)}_{\text{V}_\ell,m}\right)\!\!=\!\!
\text{p}\left(\random{d}^{(SH/m)}_{\text{V}_\ell,t_\text{pred},\text{lon}}
\vert h^{(MH)}_{\text{V}_\ell,m}\right)\nonumber\\
& \quad \quad\quad\quad \quad \quad\quad\quad \cdot
\text{p}\left(\random{d}^{(SH/m)}_{\text{V}_\ell,t_\text{pred},\text{lat}}
\vert h^{(MH)}_{\text{V}_\ell,m}\right).
\label{EqJointPdfDeviationsLat}
\end{align}

In order to implement the approach, a quantization of the
deviations
$\random{d}^{(SH/m)}_{\text{V}_\ell,t_\text{pred},\text{lon}}$ and
$\random{d}^{(SH/m)}_{\text{V}_\ell,t_\text{pred},\text{lat}}$
into totally $N$ discrete values is performed and the probability
of each outcome is computed based on
Eq.~(\ref{EqJointPdfDeviationsLat}). Thus the total number of
hypotheses possible for a vehicle $\text{V}_\ell$ at time instance
$t_\text{pred}$ will be equal to $S$, which is $M$ \emph{main
hypotheses} times $N$ \emph{sub hypotheses}.

This way, taking into account the \emph{main} and \emph{sub
hypotheses}, it is possible to assign probabilities to $S$
trajectories and to construct the $S$ dimensional vector
$\textbf{p}(h_{{\text{V}_\ell},t_\text{pred}})$. The vector holds
the probabilities for all $S$ multiple hypotheses of the traffic
object $\text{V}_\ell$ at $t_\text{pred}$.
\subsection{Construction of POGs}
\label{subsec:PredictedGrid}
\begin{figure}[htb!]
    \begin{scriptsize}
    \vspace{-0.5cm}
    \centering
    \def\svgwidth{0.33\textwidth}
    \centerline{\import{Figures/}{PoGScenario.pdf_tex}}
    \end{scriptsize}
    \vspace{-0.3cm}
\def\figurename{Figure}
\caption{Predicted-Occupancy Grid}%
\vspace{-0.2cm}%
\label{FigPredictedOccupancyGrid}
\end{figure}
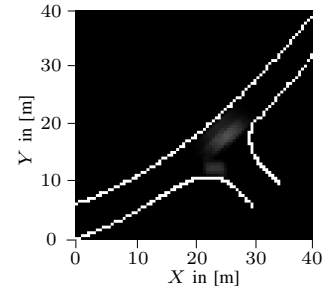

The estimation of
$\textbf{p}(h_{{\text{V}_\ell},t_\text{pred}})$ is followed by the
generation of a probabilistic representation of the future traffic
scene termed as the POG. The area belonging to a traffic scenario
is divided into cells of length $\ell_\text{cell}$ and width
$w_\text{cell}$ leading to $I$ columns and $J$ rows in
the grid. A POG $\mathcal{G}_{t_\text{pred}}$ is computed for each
prediction time instance $t_\text{pred}$. For a given prediction
horizon, there are $\kappa$ POGs, one for each prediction time.
Let the $(i,j)\text{-th}$ cell of the POG at a given prediction
instance $t_\text{pred}$ be denoted as $g_{t_\text{pred}}^{ij}$.
It is important to note that a cell of a POG can be occupied
simultaneously by multiple hypotheses of multiple traffic
participants at a given prediction instance. The probability of
occupancy $\text{p}(o_{t_\text{pred}}^{ij})$ of the cell
$g_{t_\text{pred}}^{ij}$ is assigned based on
$\textbf{p}(h_{{\text{V}_\ell},t_\text{pred}})$. Let
$\boldsymbol{r}_{\text{V}_\ell,t_\text{pred}}^{ij}\in\{0,1\}^S$ be a
binary vector of size  $S$. The $s$-th element
${r}_{\text{V}_\ell,t_\text{pred},s}^{ij}$ denotes the occupancy
of the $s$-th hypothesis of the traffic object $\text{V}_\ell$ in
the $(i,j)\text{-th}$ cell at the instance $t_\text{pred}$
\[
 {r}_{\text{V}_\ell,t_\text{pred},s}^{ij}\!\!=\!\!
  \begin{cases}
   1 & \text{if } (i,j)\text{-th cell occupied by hypothesis}~s \\& \text{of the}~\ell\text{-th object at instance}~t_\text{pred}\\
   0 & \text{if } (i,j)\text{-th cell unoccupied by hypothesis}~s \\& \text{of the}~\ell\text{-th object at instance}~t_\text{pred}.
  \end{cases}
\]
The probability of occupancy of the $(i,j)\text{-th}$ cell in
$\mathcal{G}_{t_\text{pred}}$ with $L$ traffic participants is
given by
\begin{align}
\text{p}(o_{t_\text{pred}}^{ij})=
\text{min}\left(1,\sum_{\ell=1}^{L}\left(\left(\boldsymbol{r}_{\text{V}_\ell,t_\text{pred}}^{ij}\right)^T\textbf{p}(h_{{\text{V}_\ell},t_\text{pred}})\right)\right).
\label{EqProbCellPOG}
\end{align}
The min-operator in Eq.~(\ref{EqProbCellPOG}) is necessary since
more objects can move into the same cell, this representing a
crash. The probability is assigned in a similar method to all the
cells, thereby generating POG $\mathcal{G}_{t_\text{pred}}$.
Fig.~\ref{FigPredictedOccupancyGrid} shows the constructed POG of
the scenario depicted in Fig.~\ref{FigMultipleHypothesis} for
$t_\text{pred}=2$\,s. The cells concerning the road information
are depicted with the probability of occupancy $1$.

\section{Predicted-Occupancy Grids Based on Machine Learning}
\label{sec:Machine Learning} The estimation of all multiple
hypotheses of traffic participants followed by the
construction of POGs using the model-based approach involves a
huge amount of computational effort. In order to enable real-time
computation of the POGs, a machine-learning approach is adopted in
this work. The goal is to perform a mapping from a suitable
representation of the current state, i.\,e., $t_\text{pred}=0$\,s,
of a traffic scenario to the $\kappa$ POGs
$\mathcal{G}_{t_\text{pred}}$. The current state representation
can be constructed in vehicles from the information provided by
exteroceptive and inertial sensors. The following subsection deals
with the generation of a suitable current state representation for
the machine-learning approach.

\subsection{Augmented Occupancy Grids}%
\label{subsec:OccGridAugment} In this work a current state
representation is proposed which consists of augmented cells in an
occupancy grid $\mathcal{OG}_0$. The grid is chosen to be equal in
size as the POGs but the values stored per cell are not estimated
probabilities but a set of attributes which describe the traffic
scenario corresponding to the cell. This representation
$\mathcal{OG}_0$ is called Augmented Occupancy Grid and the
attributes per cell are: the occupancy, the velocity, the
orientation, the longitudinal and lateral acceleration of the
traffic object occupying it. For example if a cell in
$\mathcal{OG}_0$ is occupied by a vehicle $V_\ell$ moving with
velocity $v_\ell$, orientation $\psi_\ell$, longitudinal
acceleration $a_{x,\ell}$ and lateral acceleration $a_{y,\ell}$
the attributes for this cell are $[1, v_\ell, \psi_\ell,
a_{x,\ell}, a_{y,\ell}]^{\text{T}}$. The information about the
road limits is also incorporated with the corresponding cells
having the attributes $[1, 0, 0, 0, 0]^{\text{T}}$. With this
notation the goal of the machine-learning algorithm is to perform
the mappings
\begin{align}
\mathcal{OG}_0 \mapsto \mathcal{G}_{t_\text{pred}}.
\label{EqMappingML}
\end{align}

\subsection{Estimation of POGs Using the RF
Algorithm}\label{SecPOGwithRF}

\subsubsection{The Random Forest Algorithm}
The mapping from Eq.~(\ref{EqMappingML}) is implemented in this
work using the RF algorithm. It is one of the most powerful
off-the-shelf machine learning algorithms. The RF algorithm has
been introduced by Breiman in~\cite{breiman01random}. It is a
randomized and aggregated version of the
well-known~\cite{breiman84} \emph{Classification And Regression
Tree} (CART) algorithm strengthened by the bagging (stands for
\emph{b}ootstrap \emph{agg}regating) technique. Given a set of
input vectors and the corresponding targets, the idea underlying
the RF algorithm is to construct a large number of simple
classifiers with low bias, e.\,g., full grown decision trees and
then to take a majority vote among the individual classifiers. It
is proven in~\cite{breiman01random} that the algorithm does not
overfit as more trees are added to the RF. The step of taking the
majority vote reduces the variance of the RF classifier without
increasing its bias compared to the bias and variance of the
individual classifiers in the ensemble~\cite{geu06}. Thus, one
obtains a classifier with low bias and low variance leading to a
small generalization error. It is interesting to note that for
classification the generalization error does not decompose into a
sum of bias and variance as it is the case in regression, but in a
fraction term containing the bias and variance, such that it is
possible to reach the minimum of the generalization error by only
decreasing the variance~\cite{fried97}. This explains why ensemble
methods like AdaBoost or RF perform so well in classification.

In order to decrease the variance term it is important that the
individual classifiers differ from each other, i.\,e., have a low
correlation. To achieve this goal a randomization source is
introduced in the construction of each tree. Although there are
many possibilities to introduce randomness, the bagging technique
is used since it allows to compute the oob-estimation of the
generalization error~\cite{breiman01random}. In~\cite{brei96b} Breiman gives empirical evidence that the oob-estimate is as accurate as using a
test set of the same size as the training set.

\subsubsection{Estimation of POGs}
\begin{figure}[htb!]
    \begin{scriptsize}
    \vspace{-0.0cm}
    \centering
    \def\svgwidth{0.35\textwidth}
    \centerline{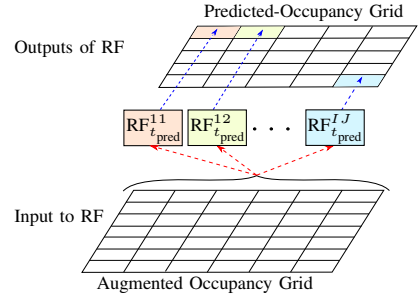}
    \end{scriptsize}
    \vspace{-0.3cm}
\def\figurename{Figure}
\caption{Per cell prediction using Random Forest}%
\vspace{-0.5cm}%
\label{FigEnsembleOfRF}
\end{figure}

The probabilities for each cell in the POG are estimated in this
work independently by using one RF classifier per cell in
$\mathcal{G}_{t_\text{pred}}$. In order to realize the mapping
from Eq.~(\ref{EqMappingML}) and with the aim to determine the likeliness of occupancy, the probability $\text{p}(o_{t_\text{pred}}^{ij})$ which has
continuous values between $0$ and $1$ is quantized into discrete
values. This quantization is explained further in
Section~\ref{subsec:GenerationOfData}. A pictorial representation of the algorithm can be seen in Fig.~\ref{FigEnsembleOfRF}. The occupancy
probability $\text{p}(o_{t_\text{pred}}^{ij})$ is estimated by the
RF classifier $\text{RF}_{t_\text{pred}}^{ij}$. The input
$\mathcal{OG}_0$ for all RF classifiers is the same. This holds
for all $\kappa$ POGs $\mathcal{G}_{t_\text{pred}}$.

\begin{figure}[htb!]
    \begin{center}
    \begin{scriptsize}
    \vspace{-0.8cm}
    \def\svgwidth{0.4\textwidth}
    \centerline{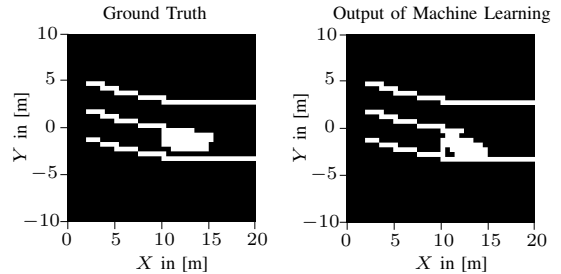}
    \end{scriptsize}
    \end{center}
    \vspace{-1.6cm}
\def\figurename{Figure}
\caption{Proof of concept}%
\vspace{-0.2cm}%
\label{FigProofOfConcept}
\end{figure}

To evaluate the ability of a set of RF classifiers to perform a
mapping as shown in Fig.\ref{FigEnsembleOfRF} in a first step a
simplified ``proof of concept'' is implemented. A traffic scenario
with a single vehicle driving on a road is considered. Given
$\mathcal{OG}_0$ the task of the set of RF classifiers is to
predict the location of the vehicle in $t_\text{pred}=1$\,s.
However, multiple hypotheses dealing with the uncertainties of the
driver behavior are not considered for this proof of concept,
leading to target values of $0$ or $1$. The generated training
data set comprises $300$ traffic scenarios.
The outputs for a test scenario and the ground truth computed by
the model-based approach can be seen in the
Fig.~\ref{FigProofOfConcept}. The Augmented Occupancy Grid
$\mathcal{OG}_0$ consists of the road limits, the middle lane
marking and a vehicle which whose center of gravity is located at
the position $(2.5\,\text{m},0\,\text{m})$. The results of this
proof of concept show the ability of a classification system as
presented in Fig.~\ref{FigEnsembleOfRF} to implement the mapping
from Eq.~(\ref{EqMappingML}).

\section{Evaluation of the Machine Learning Approach}
\label{sec:Evaluation} This section explains the generation of
data for training and testing the machine-learning approach
followed by the simulation results with their corresponding
quality assessment.
\subsection{Generation of Data}
\label{subsec:GenerationOfData} 
\begin{figure}[htb!]
    \begin{scriptsize}
    \vspace{0.05cm}
    \centering
    \def\svgwidth{0.33\textwidth}
    \centerline{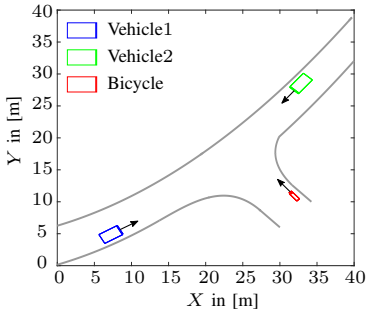}
    \end{scriptsize}
    \vspace{-0.2cm}
\def\figurename{Figure}
\caption{Scenario under consideration}%
\vspace{-0.6cm}%
\label{FigScenario}
\end{figure}

The input and the target of the
training set for the machine learning algorithm is generated using
the methods explained in Subsection~\ref{subsec:OccGridAugment}
and Section~\ref{sec:Predicted-Occupancy}, respectively. A complex
traffic scenario with an intersection on a curved road and with
$3$ traffic participants ($2$ vehicles and $1$ bicycle) on a span
of $40 \times 40$ meters is considered. The traffic scene is
visualized in Fig.~\ref{FigScenario}.
Parameters used for the generation of the data are:
$\ell_\text{cell}=w_\text{cell}=0.5$\,m,
$a_\text{decel,max}=9$\,m/s$^2$,
$a_\text{accel,max}=4.5$\,m/s$^2$, and
$a_\text{lat,max}=7$\,m/s$^2$. Thus, a POG
$\mathcal{G}_{t_\text{pred}}$ has $80 \times 80$ cells and leads
to a total number of $6400$ RF classifiers.

A number of $972$ initial traffic scenarios represented by $972$
Augmented Occupancy Grids $\mathcal{OG}_0$ are generated by
varying the parameters of the traffic participants such as their
position, velocity and acceleration. The position for vehicles has
been varied over a range of $10$\,m, the velocity over a range of
$20$\,km/h and the longitudinal acceleration over $2.5$\,m/s$^2$.
The position for the bicycle has been varied over a range of
$6$\,m, the velocity over a range of $10$\,km/h and the
longitudinal acceleration over $1$\,m/s$^2$.

The target data, i.\,e., the POGs $\mathcal{G}_{t_\text{pred}}$
are obtained using the model-based approach from
Subsection~\ref{subsec:PredictedGrid}. The prediction horizon is
set to be $2$\,s and is represented by $\kappa=3$ POGs
$\mathcal{G}_{t_\text{pred}}$, with $t_\text{pred}$ being
$0.5$\,s, $1.0$\,s and $2.0$\,s.

As mentioned in Subsection~\ref{SecPOGwithRF} the continuous
values $\text{p}(o_{t_\text{pred}}^{ij})$ must be quantized in
order to estimate the likeliness of occupancy. The quantized
probability value is denoted as
$\text{p}_{q}(o_{t_\text{pred}}^{ij})$.  The quantization strategy
adopted in this work is
\begin{align}
\text{p}_{q}(o_{t_\text{pred}}^{ij})\in\{0, 0.25, 0.5, 0.75,
1.0\}.
\end{align}
These quantization values represent the degree of likeliness of
occupancy. For example, $0.25$ would represent less likeliness and
$0.75$ would state that it is more likely for the corresponding
cell to be occupied.

As mentioned above the number of traffic scenarios is $972$.
Two-thirds of the scenarios are used for training and the
remaining is used as a validation set. With $\kappa=3$ the
prediction for $0.5$\,s, $1.0$\,s and $2.0$\,s leads to $3\times
6400$ RF classifiers, each having $700$ training scenarios.

\subsection{Quality Assessment}
It is important to quantify the quality of the machine-learning
approach adopted for determining the probabilities of POG. With
the oob-estimate of one RF classifier as introduced in
Subsection~\ref{SecPOGwithRF}, the quality assessment can only be
computed per cell of the POG. Since it is necessary to find the
overall quality of the POG estimate 
$\hat{\mathcal{G}}_{t_\text{pred}}$ computed using the
machine-learning approach, a new quality quantity is formulated.
Considering the fact that most of the cells in a POG have an
occupancy probability of $0$, such as regions outside the roads,
it is useful to define region of interest while estimating the
quality of prediction. Thus, only the non-zero cells are
considered for the quality assessment. Let $\mathcal{B}$ denote
the set of cells with a non-zero value in the estimated POG
$\hat{\mathcal{G}}_{t_\text{pred}}$ based on the machine-learning
approach. Let $\mathcal{D}$ denote the set of cells with a
non-zero value in the corresponding ground truth POG
$\mathcal{G}_{t_\text{pred}}$ from the model-based approach. With
$K$ representing the cardinality of the set
$\left(\mathcal{B}\cup\mathcal{D}\right)\setminus
\left(\mathcal{B}\cap\mathcal{D}\right)$,
\begin{align}
K=\left\vert \left(\mathcal{B}\cup\mathcal{D}\right)\setminus
\left(\mathcal{B}\cap\mathcal{D}\right) \right\vert,
\end{align}
the quality measure is defined as
\begin{align}
\label{Eq:MSE}
\epsilon_{{t_\text{pred}}}=\sqrt{\frac{1}{K}\sum\limits_{i=1}^{I}\sum\limits_{j=1}^{J}\Big(\hat{\text{p}}(o_{t_\text{pred}}^{ij})-\text{p}_{q}(o_{t_\text{pred}}^{ij})\Big)^2},
\end{align}
where $\hat{\text{p}}(o_{t_\text{pred}}^{ij})$ is the estimated
value of the $(i,j)$-th cell from the machine-learning approach
and $\text{p}_{q}(o_{t_\text{pred}}^{ij})$ is the quantized true
occupancy probability of $(i,j)$-th cell from the model-based
approach.

\subsection{Simulation Results}\label{SubSecSimRes}
\begin{figure}[htb!]
    \begin{scriptsize}
    \vspace{-0.0cm}
    \centering
    \def\svgwidth{0.45\textwidth}
    \centerline{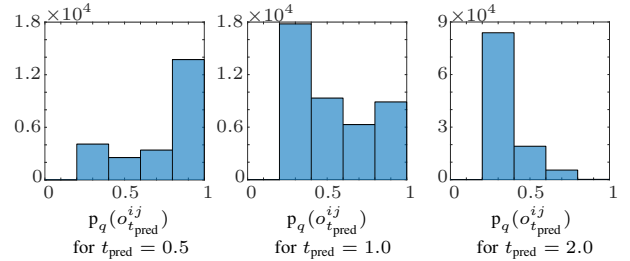}
    \end{scriptsize}
    \vspace{-0.1cm}
\def\figurename{Figure}
\caption{Histograms of $\text{p}_{q}(o_{t_\text{pred}}^{ij})$ for varying $t_\text{pred}$ }%
\vspace{-0.3cm}%
\label{FigProbOfOcc}
\end{figure}

The simulation results for prediction time instances
$t_\text{pred}$ of $0.5$\,s, $1.0$\,s and $2.0$\,s are presented
here. The histograms of non-zero
$\text{p}_{q}(o_{t_\text{pred}}^{ij})$ for $t_\text{pred}$ of
$0.5$\,s, $1.0$\,s and $2.0$\,s over the $272$ test scenarios are
shown in the Fig.~\ref{FigProbOfOcc}. It can be seen that the
number of cells taking lower values of
$\text{p}_{q}(o_{t_\text{pred}}^{ij})$ increases as
$t_\text{pred}$ increases. This reflects the fact that the
uncertainties regarding the traffic objects increase with an
increase in the prediction horizon.

In order to perform a detailed quality assessment, error estimates
according to Eq.~(\ref{Eq:MSE}) for low, middle and high values of
$\text{p}_{q}(o_{t_\text{pred}}^{ij})$ are computed, leading to
the three values $\epsilon_{{t_\text{pred}},\text{low}}$,
$\epsilon_{{t_\text{pred}},\text{med}}$ and
$\epsilon_{{t_\text{pred}},\text{high}}$ for each $t_\text{pred}$.
The occupancy probability value corresponding to
$\epsilon_{{t_\text{pred}},\text{low}}$ is
$\text{p}_{q}(o_{t_\text{pred}}^{ij})=0.25$. The occupancy
probability values corresponding to
$\epsilon_{{t_\text{pred}},\text{med}}$ are
$\text{p}_{q}(o_{t_\text{pred}}^{ij})=0.5$ and
$\text{p}_{q}(o_{t_\text{pred}}^{ij})=0.75$. The occupancy
probability value corresponding to
$\epsilon_{{t_\text{pred}},\text{high}}$ is
$\text{p}_{q}(o_{t_\text{pred}}^{ij})=1.0$.

\begin{figure}[htb!]
    \begin{scriptsize}
    \vspace{-1.5cm}
    \centering
    \def\svgwidth{0.45\textwidth}
    \centerline{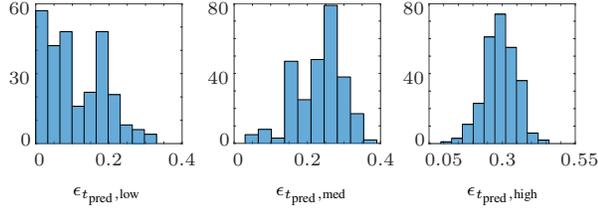}
    \end{scriptsize}
    \vspace{-1.5cm}
\def\figurename{Figure}
\caption{Histograms of ${\epsilon}_{{{t_\text{pred}}}}$ for $t_\text{pred}=0.5$}%
\vspace{-0.0cm}%
\label{FigHistOfError05}
\end{figure}

\begin{figure}[htb!]
    \begin{scriptsize}
    \vspace{-2.2cm}
    \centering
    \def\svgwidth{0.45\textwidth}
    \centerline{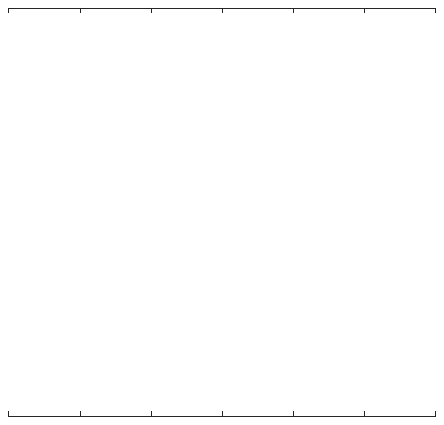}
    \end{scriptsize}
    \vspace{-1.5cm}
\def\figurename{Figure}
\caption{Histograms of ${\epsilon}_{{{t_\text{pred}}}}$ for $t_\text{pred}=1.0$}%
\vspace{-0.2cm}%
\label{FigHistOfError}
\end{figure}

\begin{figure}[htb!]
    \begin{scriptsize}
    \vspace{-1.7cm}
    \centering
    \def\svgwidth{0.45\textwidth}
    \centerline{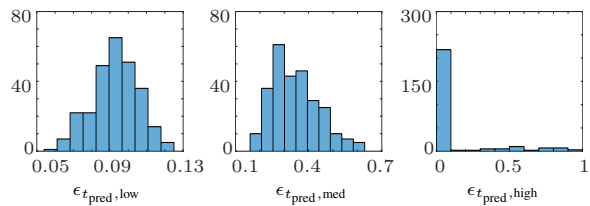}
    \end{scriptsize}
    \vspace{-1.5cm}
\def\figurename{Figure}
\caption{Histograms of ${\epsilon}_{{{t_\text{pred}}}}$ for $t_\text{pred}=2.0$}%
\vspace{-0.2cm}%
\label{FigHistOfError20}
\end{figure}

The estimated mean errors
$\bar{\epsilon}_{{t_\text{pred}},\text{low}}$,
$\bar{\epsilon}_{{t_\text{pred}},\text{med}}$ and
$\bar{\epsilon}_{{t_\text{pred}},\text{high}}$ over all the test
scenarios for varying $t_\text{pred}$ are presented in
Table~\ref{Tab:Error}. The corresponding histograms are visualized
in Fig.~\ref{FigHistOfError05}, Fig.~\ref{FigHistOfError} and
Fig.~\ref{FigHistOfError20}. As it can be seen the main outcome
can be stated as: low occupancy probabilities are predicted by the
machine learning approach as low probabilities, middle-valued
occupancy probabilities are predicted as middle-valued
probabilities and high occupancy probabilities are predicted as
high probabilities, independently of the prediction horizon. The
machine learning approach tends to overestimate low occupancy
probabilities with $\approx 10$\,$\%$ but the estimates never
reach high occupancy probabilities. For high occupancy
probabilities the machine-learning approach underestimates the
occupancy probability with $\approx 10 - 30$\,$\%$ depending on
the prediction horizon, but only in a single scenario for
$t_\text{pred}=2.0$\,s the estimate is associated to an occupancy
probability of zero. However, it must be taken into account that
an occupancy probability of $1$ is very unlikely for
$t_\text{pred}=2.0$\,s as presented in Fig.~\ref{FigProbOfOcc}.

%
%
%
%

\begin{table}[!t]
\vspace{0.2cm}
\renewcommand{\arraystretch}{1.3}
\caption{Comparison of the errors for varying prediction times }
\label{Tab:Error} \centering
\begin{tabular}{c c c c}
\hline
 & $t_\text{pred}=0.5$ & $t_\text{pred}=1.0$ & $t_\text{pred}=2.0$ \\
\hline
$\bar{\epsilon}_{{{t_\text{pred}},\text{low}}}$ & $0.1090$ & $0.1365$ & $0.0930$\\
\hline
$\bar{\epsilon}_{{{t_\text{pred}},\text{med}}}$ & $0.2937$ & $0.3653$ & $0.3488$\\
\hline
$\bar{\epsilon}_{{{t_\text{pred}},\text{high}}}$ & $0.2315$ & $0.3171$ & $0.1071$\\
\hline
\end{tabular}
\vspace{-0.5cm}
\end{table}

\begin{figure}[htb!]
    \begin{scriptsize}
    \vspace{-1.0cm}
    \centering
    \def\svgwidth{0.47\textwidth}
    \centerline{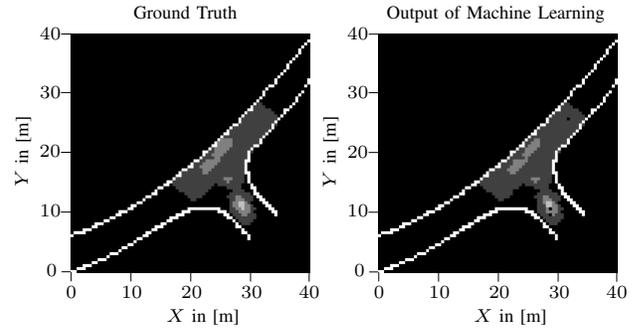}
    \end{scriptsize}
    \vspace{-0.9cm}
\def\figurename{Figure}
\caption{Comparison between ground truth and estimated Predicted-Occupancy Grid for $t_\text{pred}=2.0$\,s}%
\vspace{-0.3cm}%
\label{FigPredicted2Sec}
\end{figure}

It is important to note that the quality measure defined in this
work by Eq.~(\ref{Eq:MSE}) does not take into account all those
cells where the classifier correctly estimates the value of
$\text{p}_{q}(o_{t_\text{pred}}^{ij})=0$. However, the correct
prediction of $\text{p}_{q}(o_{t_\text{pred}}^{ij})=0$ is also a
significant part of the classification task. For example, the mean
error per cell for the POG ${\mathcal{G}}_{t_\text{pred}}$ shown
in the Fig.~\ref{FigPredicted2Sec} is $0.0024$, whereas
$\epsilon_{{t_\text{pred}}}$ has the value $0.1094$. So, the
quality assessment defined in this work does not reward the
correct prediction of free space, thus leading to larger values
than the mean error per cell.



A key benefit of this work can be visualized exemplarily using
Fig.~\ref{FigPredicted2Sec}. It shows the quantized values of a
POG $\mathcal{G}_{t_\text{pred}}$ and the corresponding
machine-learning estimate $\hat{\mathcal{G}}_{t_\text{pred}}$ for
$t_\text{pred}=2.0$\,s. The initial state of the corresponding
traffic scenario is shown in the Fig.~\ref{FigScenario}. As it can
be seen, the machine learning approach is able to generate good
estimates of POGs. The main advantage of using the machine
learning approach is that the estimate
$\hat{\mathcal{G}}_{t_\text{pred}}$ can be generated with
relatively low computational effort as soon as the set of trained
RF classifiers already exists. The computational effort for
generating $\mathcal{G}_{t_\text{pred}}$ using the model-based
approach is huge and can only be done offline. Measurements in
Matlab show that the computational time of the machine-learning
approach is approx. $5$ times faster than the model-based
approach. Taking into account that the presented machine learning
approach is highly parallelizable a real-time computation, for
example on GPUs, is realistic.


\section{Applications in Vehicle Safety}
\label{sec:Applications} The computation of POGs in real time, for
example using the machine learning approach, would enable the
improvement of numerous applications in the field of vehicle
safety. Some of them are presented in the following.

\subsection{Trajectory Planning}
Vehicle safety systems that are performing autonomous
interventions in the vehicle dynamics, in order to avoid a
collision or to mitigate the outcome of a collision, have to plan
a desired trajectory and then to control the vehicle according to
this trajectory. The planning of trajectories is a challenging
task and current approaches to solve this problem in real-time
rely on probabilistic sampling methods like the Rapidly-exploring
Random Tree (RRT) algorithm~\cite{LaValle98}. An extension of the
RRT called Augmented-Closed-Loop-RRT is introduced
in~\cite{chaulwar16} for applications with multiple dynamic
objects in complex traffic scenarios. The latter algorithm takes
into account where objects will be located in future time
instances in order to find collision-free trajectories. POGs that
consider multiple hypotheses of objects can be used to improve
this algorithm by searching for trajectories with a very low risk
of collision. This can be realized by choosing trajectories which
pass only through POG cells with very low probability of being
occupied by other objects. Since the POGs can be implemented
efficiently using the machine learning approach, the entire
algorithm will maintain its fast computational speed.

\subsection{Criticality Estimation}
POGs can also be used to estimate the risk of traffic scenarios. A
survey on risk-quantities for road traffic is presented
in~\cite{Lefevre14}. A novel and natural way to describe the risk
of a traffic scenario can be introduced using POGs. Considering a
prediction time horizon of $t_\text{pred}$ and $\kappa$ POGs
$\mathcal{G}_{t_\text{pred}}$, also a number of $\kappa$
criticality-values $c_{t_\text{pred}}$ can be computed, one for
each prediction time. If the $(i,j)\text{-th}$ cell in the POG
$\mathcal{G}_{t_\text{pred}}$ is denoted with
$g_{t_\text{pred}}^{ij}$ then the criticality
$c_{t_\text{pred}}^{ij}$ can be obtained by estimating in a first
step the probability of $g_{t_\text{pred}}^{ij}$ being occupied by
another object than the EGO-vehicle, which is the vehicle in which
the safety systems operate. Using a POG based on machine-learning
as presented above, this value can be computed directly with a set
of RF classifiers. Then the probability of the cell being occupied
by the EGO-vehicle can be computed again using the set of RF
classifiers. With these two estimated probabilities the
criticality $c_{t_\text{pred}}^{ij}$  for the $(i,j)\text{-th}$
cell can be obtained by multiplying them, assuming statistical
independence. The criticality $c_{t_\text{pred}}$ for the
prediction time instance $t_\text{pred}$ can be defined to be the
maximum among the criticality values in the cells
\begin{align}
c_{t_\text{pred}}=\underset{i,j}{\text{max}}\{c_{t_\text{pred}}^{ij}\}.
\end{align}
\quad \, \quad   \vspace{-0.4cm}\\%
The criticality $c_\text{total}$ of the scenario can be defined to
be the maximum among the $\kappa$ values $c_{t_\text{pred}}$.
Such a criticality measure can be used in vehicle safety systems
to trigger the activation of actuators.

\subsection{Clustering of Scenarios for Testing}
The testing of vehicle safety systems is a crucial and costly task
in the development process. An important aspect in the testing
procedure is the choice of the test-scenarios. They must be chosen
to show that the system has a very high true-positive rate and a
very low false-positive rate. It depends on the system under
consideration which test-scenarios are best suited to estimate
these rates accurately. But a common goal for the testing of all
systems is the reduction of the number of tests to a
representative set. One approach to achieve this reduction is to
use clustering techniques. POGs offer the possibility to cluster
scenarios by taking into account also hypotheses about the future
development of a traffic scenario. Then, one cluster could be
represented by a single Augmented Occupancy Grid
$\mathcal{OG}_0$.


\section{Conclusions and Future Work}
\label{sec:Conclusions} This paper presents novel methods for the
representation of the current traffic scenario and for the
prediction of the future traffic scenario using occupancy grids.
The prediction and its representation takes into account the
uncertainties regarding the motion of traffic participants. A
model-based approach and a machine learning approach using the
Random Forest classifier are introduced for the prediction task.
Using simulations, it is shown that the machine learning approach
is able to estimate the future traffic scenario well. With trained
classifiers, the computational effort to perform predictions is
much smaller than using the model-based approach. This could
enable the real-time computation of Predicted Occupancy Grids.

Future work will focus on training a set of Random Forest
classifiers for scenarios with varying number of objects and
varying road infrastructures. Additionally, Predicted Occupancy
Grids will be included in vehicle safety algorithms for trajectory
planning and criticality estimation.

%

\addtolength{\textheight}{-12cm}   



\end{document}

%% file: Figures/TwoTrackModel.pdf_tex
\begingroup%
  \makeatletter%
  \providecommand\color[2][]{%
    \errmessage{(Inkscape) Color is used for the text in Inkscape, but the package 'color.sty' is not loaded}%
    \renewcommand\color[2][]{}%
  }%
  \providecommand\transparent[1]{%
    \errmessage{(Inkscape) Transparency is used (non-zero) for the text in Inkscape, but the package 'transparent.sty' is not loaded}%
    \renewcommand\transparent[1]{}%
  }%
  \providecommand\rotatebox[2]{#2}%
  \ifx\svgwidth\undefined%
    \setlength{\unitlength}{841.88976378bp}%
    \ifx\svgscale\undefined%
      \relax%
    \else%
      \setlength{\unitlength}{\unitlength * \real{\svgscale}}%
    \fi%
  \else%
    \setlength{\unitlength}{\svgwidth}%
  \fi%
  \global\let\svgwidth\undefined%
  \global\let\svgscale\undefined%
  \makeatother%
  \begin{picture}(1,0.70707071)%
    \put(0,0){\includegraphics[width=\unitlength,page=1]{TwoTrackModel.pdf}}%
    \put(0.05828738,0.39742101){\color[rgb]{0,0,0}\rotatebox{86.26285209}{\makebox(0,0)[lb]{\smash{$w$}}}}%
    \put(0,0){\includegraphics[width=\unitlength,page=2]{TwoTrackModel.pdf}}%
    \put(0.14116706,0.45071484){\color[rgb]{0,0,0}\rotatebox{90.31014251}{\makebox(0,0)[lb]{\smash{$\frac{w}{2}$}}}}%
    \put(0,0){\includegraphics[width=\unitlength,page=3]{TwoTrackModel.pdf}}%
    \put(0.89817155,0.56984838){\color[rgb]{0,0,0}\makebox(0,0)[lb]{\smash{$\delta_\mathrm{fl}$}}}%
    \put(0.89410134,0.29322788){\color[rgb]{0,0,0}\makebox(0,0)[lb]{\smash{$\delta_\mathrm{fr}$}}}%
    \put(0.37428766,0.56620805){\color[rgb]{0,0,0}\makebox(0,0)[lb]{\smash{$\delta_\mathrm{rl}$}}}%
    \put(0.37062751,0.2801905){\color[rgb]{0,0,0}\makebox(0,0)[lb]{\smash{$\delta_\mathrm{rr}$}}}%
    \put(0,0){\includegraphics[width=\unitlength,page=4]{TwoTrackModel.pdf}}%
    \put(0.52908616,0.46817654){\color[rgb]{0,0,0}\makebox(0,0)[lb]{\smash{$\boldsymbol{v}$}}}%
    \put(0,0){\includegraphics[width=\unitlength,page=5]{TwoTrackModel.pdf}}%
    \put(0.54772508,0.42000956){\color[rgb]{0,0,0}\makebox(0,0)[lb]{\smash{$\beta$}}}%
    \put(0.06776955,0.72998577){\color[rgb]{0,0,0}\makebox(0,0)[lb]{\smash{}}}%
    \put(0,0){\includegraphics[width=\unitlength,page=6]{TwoTrackModel.pdf}}%
    \put(0.28567404,0.18688403){\color[rgb]{0,0,0}\makebox(0,0)[lb]{\smash{$\ell_\mathrm{r}$}}}%
    \put(0.55704793,0.18235957){\color[rgb]{0,0,0}\makebox(0,0)[lb]{\smash{$\ell_\mathrm{f}$}}}%
    \put(0.46428235,0.13395039){\color[rgb]{0,0,0}\makebox(0,0)[lb]{\smash{$\ell$}}}%
    \put(0.41859204,0.431209){\color[rgb]{0,0,0}\makebox(0,0)[lb]{\smash{C}}}%
    \put(0.3218096,0.5126505){\color[rgb]{0,0,0}\makebox(0,0)[lb]{\smash{$F_{x\mathrm{rl}}$}}}%
    \put(0.31474983,0.23363754){\color[rgb]{0,0,0}\makebox(0,0)[lb]{\smash{$F_{y\mathrm{rr}}$}}}%
    \put(0.22752244,0.33770187){\color[rgb]{0,0,0}\makebox(0,0)[lb]{\smash{$F_{y\mathrm{rr}}$}}}%
    \put(0.22627477,0.61671465){\color[rgb]{0,0,0}\makebox(0,0)[lb]{\smash{$F_{y\mathrm{rl}}$}}}%
    \put(0.89817155,0.2334715){\color[rgb]{0,0,0}\makebox(0,0)[lb]{\smash{$F_{x\mathrm{fr}}$}}}%
    \put(0.80810295,0.35961287){\color[rgb]{0,0,0}\makebox(0,0)[lb]{\smash{$F_{y\mathrm{fr}}$}}}%
    \put(0.80381402,0.63992694){\color[rgb]{0,0,0}\makebox(0,0)[lb]{\smash{$F_{y\mathrm{fl}}$}}}%
    \put(0.90246054,0.5137856){\color[rgb]{0,0,0}\makebox(0,0)[lb]{\smash{$F_{x\mathrm{fl}}$}}}%
    \put(0,0){\includegraphics[width=\unitlength,page=7]{TwoTrackModel.pdf}}%
    \put(0.38349442,0.36428473){\color[rgb]{0,0,0}\makebox(0,0)[lb]{\smash{$\dot{\psi}$}}}%
    \put(0,0){\includegraphics[width=\unitlength,page=8]{TwoTrackModel.pdf}}%
    \put(0.12847293,0.09616117){\color[rgb]{0,0,0}\makebox(0,0)[lb]{\smash{$\boldsymbol{e}_x$}}}%
    \put(0.05093742,0.06735842){\color[rgb]{0,0,0}\makebox(0,0)[lb]{\smash{$\boldsymbol{e}_z$}}}%
    \put(0.06705707,0.16334253){\color[rgb]{0,0,0}\makebox(0,0)[lb]{\smash{$\boldsymbol{e}_y$}}}%
    \put(-0.09502432,0.9882529){\color[rgb]{0,0,0}\makebox(0,0)[lt]{\begin{minipage}{0.64616536\unitlength}\raggedright \end{minipage}}}%
  \end{picture}%
\endgroup%

%% file: Figures/multipleHypothesis.pdf_tex
\begingroup%
  \makeatletter%
  \providecommand\color[2][]{%
    \errmessage{(Inkscape) Color is used for the text in Inkscape, but the package 'color.sty' is not loaded}%
    \renewcommand\color[2][]{}%
  }%
  \providecommand\transparent[1]{%
    \errmessage{(Inkscape) Transparency is used (non-zero) for the text in Inkscape, but the package 'transparent.sty' is not loaded}%
    \renewcommand\transparent[1]{}%
  }%
  \providecommand\rotatebox[2]{#2}%
  \newcommand*\fsize{\dimexpr\f@size pt\relax}%
  \newcommand*\lineheight[1]{\fontsize{\fsize}{#1\fsize}\selectfont}%
  \ifx\svgwidth\undefined%
    \setlength{\unitlength}{623.62204724bp}%
    \ifx\svgscale\undefined%
      \relax%
    \else%
      \setlength{\unitlength}{\unitlength * \real{\svgscale}}%
    \fi%
  \else%
    \setlength{\unitlength}{\svgwidth}%
  \fi%
  \global\let\svgwidth\undefined%
  \global\let\svgscale\undefined%
  \makeatother%
  \begin{picture}(1,0.86363636)%
    \lineheight{1}%
    \setlength\tabcolsep{0pt}%
    \put(0,0){\includegraphics[width=\unitlength,page=1]{multipleHypothesis.pdf}}%
  \end{picture}%
\endgroup%

%% file: Figures/PoGScenario.pdf_tex
\begingroup%
  \makeatletter%
  \providecommand\color[2][]{%
    \errmessage{(Inkscape) Color is used for the text in Inkscape, but the package 'color.sty' is not loaded}%
    \renewcommand\color[2][]{}%
  }%
  \providecommand\transparent[1]{%
    \errmessage{(Inkscape) Transparency is used (non-zero) for the text in Inkscape, but the package 'transparent.sty' is not loaded}%
    \renewcommand\transparent[1]{}%
  }%
  \providecommand\rotatebox[2]{#2}%
  \ifx\svgwidth\undefined%
    \setlength{\unitlength}{841.88976378bp}%
    \ifx\svgscale\undefined%
      \relax%
    \else%
      \setlength{\unitlength}{\unitlength * \real{\svgscale}}%
    \fi%
  \else%
    \setlength{\unitlength}{\svgwidth}%
  \fi%
  \global\let\svgwidth\undefined%
  \global\let\svgscale\undefined%
  \makeatother%
  \begin{picture}(1,0.70707071)%
    \put(0,0){\includegraphics[width=\unitlength,page=1]{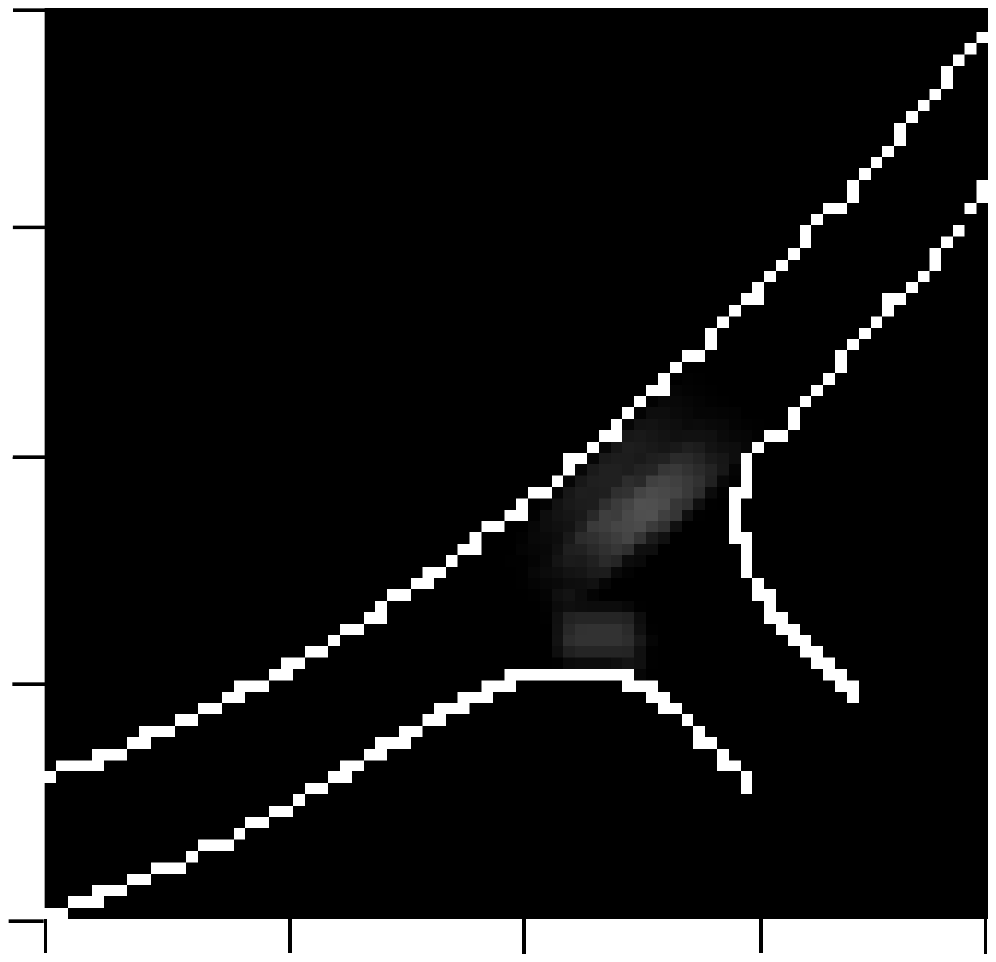}}%
    \put(0.25336624,0.07198657){\color[rgb]{0,0,0}\makebox(0,0)[lb]{\smash{$0$}}}%
    \put(0.38353866,0.07198657){\color[rgb]{0,0,0}\makebox(0,0)[lb]{\smash{$10$}}}%
    \put(0.51735596,0.07198232){\color[rgb]{0,0,0}\makebox(0,0)[lb]{\smash{$20$}}}%
    \put(0.65117302,0.07198232){\color[rgb]{0,0,0}\makebox(0,0)[lb]{\smash{$30$}}}%
    \put(0.77797654,0.07198657){\color[rgb]{0,0,0}\makebox(0,0)[lb]{\smash{$40$}}}%
    \put(0.19959591,0.11464378){\color[rgb]{0,0,0}\makebox(0,0)[lb]{\smash{$0$}}}%
    \put(0.18609292,0.24886332){\color[rgb]{0,0,0}\makebox(0,0)[lb]{\smash{$10$}}}%
    \put(0.1865707,0.37820255){\color[rgb]{0,0,0}\makebox(0,0)[lb]{\smash{$20$}}}%
    \put(0.18654622,0.51011115){\color[rgb]{0,0,0}\makebox(0,0)[lb]{\smash{$30$}}}%
    \put(0.18691375,0.63230105){\color[rgb]{0,0,0}\makebox(0,0)[lb]{\smash{$40$}}}%
    \put(0.47368491,0.02945444){\color[rgb]{0,0,0}\makebox(0,0)[lb]{\smash{$X$ in [m]}}}%
    \put(0.16583556,0.29437183){\color[rgb]{0,0,0}\rotatebox{89.78945262}{\makebox(0,0)[lb]{\smash{$Y$ in [m]}}}}%
  \end{picture}%
\endgroup%

%% file: Figures/ensembleRF.pdf_tex
\begingroup%
  \makeatletter%
  \providecommand\color[2][]{%
    \errmessage{(Inkscape) Color is used for the text in Inkscape, but the package 'color.sty' is not loaded}%
    \renewcommand\color[2][]{}%
  }%
  \providecommand\transparent[1]{%
    \errmessage{(Inkscape) Transparency is used (non-zero) for the text in Inkscape, but the package 'transparent.sty' is not loaded}%
    \renewcommand\transparent[1]{}%
  }%
  \providecommand\rotatebox[2]{#2}%
  \ifx\svgwidth\undefined%
    \setlength{\unitlength}{841.88976378bp}%
    \ifx\svgscale\undefined%
      \relax%
    \else%
      \setlength{\unitlength}{\unitlength * \real{\svgscale}}%
    \fi%
  \else%
    \setlength{\unitlength}{\svgwidth}%
  \fi%
  \global\let\svgwidth\undefined%
  \global\let\svgscale\undefined%
  \makeatother%
  \begin{picture}(1,0.70707071)%
    \put(0,0){\includegraphics[width=\unitlength,page=1]{ensembleRF.pdf}}%
    \put(0.24759052,0.055148){\color[rgb]{0,0,0}\makebox(0,0)[lb]{\smash{$\text{Augmented Occupancy Grid}$}}}%
    \put(0,0){\includegraphics[width=\unitlength,page=2]{ensembleRF.pdf}}%
    \put(0.07204771,0.55927381){\color[rgb]{0,0,0}\makebox(0,0)[lb]{\smash{$\text{Outputs of RF}$}}}%
    \put(0.07155217,0.19822531){\color[rgb]{0,0,0}\makebox(0,0)[lb]{\smash{$\text{Input to RF}$}}}%
    \put(0.47686275,0.63592568){\color[rgb]{0,0,0}\makebox(0,0)[lb]{\smash{$\text{Predicted-Occupancy Grid}$}}}%
    \put(0,0){\includegraphics[width=\unitlength,page=3]{ensembleRF.pdf}}%
    \put(0.3113298,0.39209742){\color[rgb]{0,0,0}\makebox(0,0)[lb]{\smash{$\text{RF}_{t_\text{pred}}^{11}$}}}%
    \put(0,0){\includegraphics[width=\unitlength,page=4]{ensembleRF.pdf}}%
    \put(0.44764376,0.39257666){\color[rgb]{0,0,0}\makebox(0,0)[lb]{\smash{$\text{RF}_{t_\text{pred}}^{12}$}}}%
    \put(0,0){\includegraphics[width=\unitlength,page=5]{ensembleRF.pdf}}%
    \put(0.69881999,0.39234159){\color[rgb]{0,0,0}\makebox(0,0)[lb]{\smash{$\text{RF}_{t_\text{pred}}^{IJ}$}}}%
  \end{picture}%
\endgroup%

%% file: Figures/OutputExample.pdf_tex
\begingroup%
  \makeatletter%
  \providecommand\color[2][]{%
    \errmessage{(Inkscape) Color is used for the text in Inkscape, but the package 'color.sty' is not loaded}%
    \renewcommand\color[2][]{}%
  }%
  \providecommand\transparent[1]{%
    \errmessage{(Inkscape) Transparency is used (non-zero) for the text in Inkscape, but the package 'transparent.sty' is not loaded}%
    \renewcommand\transparent[1]{}%
  }%
  \providecommand\rotatebox[2]{#2}%
  \ifx\svgwidth\undefined%
    \setlength{\unitlength}{841.88976378bp}%
    \ifx\svgscale\undefined%
      \relax%
    \else%
      \setlength{\unitlength}{\unitlength * \real{\svgscale}}%
    \fi%
  \else%
    \setlength{\unitlength}{\svgwidth}%
  \fi%
  \global\let\svgwidth\undefined%
  \global\let\svgscale\undefined%
  \makeatother%
  \begin{picture}(1,0.70707071)%
    \put(0.16215322,0.57213808){\color[rgb]{0,0,0}\makebox(0,0)[lb]{\smash{$\text{Ground Truth}$}}}%
    \put(0.59180158,0.57231787){\color[rgb]{0,0,0}\makebox(0,0)[lb]{\smash{$\text{Output of Machine Learning}$}}}%
    \put(0,0){\includegraphics[width=\unitlength,page=1]{OutputExample.pdf}}%
    \put(0.53379775,0.18397123){\color[rgb]{0,0,0}\makebox(0,0)[lb]{\smash{$-10$}}}%
    \put(0.54945302,0.27426886){\color[rgb]{0,0,0}\makebox(0,0)[lb]{\smash{$-5$}}}%
    \put(0.57611438,0.36098796){\color[rgb]{0,0,0}\makebox(0,0)[lb]{\smash{$0$}}}%
    \put(0.57611438,0.45053476){\color[rgb]{0,0,0}\makebox(0,0)[lb]{\smash{$5$}}}%
    \put(0.56045911,0.53348355){\color[rgb]{0,0,0}\makebox(0,0)[lb]{\smash{$10$}}}%
    \put(0.54067997,0.30301411){\color[rgb]{0,0,0}\rotatebox{89.76547938}{\makebox(0,0)[lb]{\smash{$Y$ in [m]}}}}%
    \put(0,0){\includegraphics[width=\unitlength,page=2]{OutputExample.pdf}}%
    \put(0.60632143,0.14909948){\color[rgb]{0,0,0}\makebox(0,0)[lb]{\smash{$0$}}}%
    \put(0.69860594,0.14909948){\color[rgb]{0,0,0}\makebox(0,0)[lb]{\smash{$5$}}}%
    \put(0.77931112,0.14909948){\color[rgb]{0,0,0}\makebox(0,0)[lb]{\smash{$10$}}}%
    \put(0.86731867,0.14909948){\color[rgb]{0,0,0}\makebox(0,0)[lb]{\smash{$15$}}}%
    \put(0.95328847,0.14909948){\color[rgb]{0,0,0}\makebox(0,0)[lb]{\smash{$20$}}}%
    \put(0.74502601,0.1049956){\color[rgb]{0,0,0}\makebox(0,0)[lb]{\smash{$X$ in [m]}}}%
    \put(0,0){\includegraphics[width=\unitlength,page=3]{OutputExample.pdf}}%
    \put(0.01158669,0.185251){\color[rgb]{0,0,0}\makebox(0,0)[lb]{\smash{$-10$}}}%
    \put(0.02724194,0.27554858){\color[rgb]{0,0,0}\makebox(0,0)[lb]{\smash{$-5$}}}%
    \put(0.05390332,0.36226768){\color[rgb]{0,0,0}\makebox(0,0)[lb]{\smash{$0$}}}%
    \put(0.05390332,0.45181448){\color[rgb]{0,0,0}\makebox(0,0)[lb]{\smash{$5$}}}%
    \put(0.03824804,0.5347633){\color[rgb]{0,0,0}\makebox(0,0)[lb]{\smash{$10$}}}%
    \put(0.01846818,0.30429381){\color[rgb]{0,0,0}\rotatebox{89.76547938}{\makebox(0,0)[lb]{\smash{$Y$ in [m]}}}}%
    \put(0,0){\includegraphics[width=\unitlength,page=4]{OutputExample.pdf}}%
    \put(0.0841106,0.15037918){\color[rgb]{0,0,0}\makebox(0,0)[lb]{\smash{$0$}}}%
    \put(0.17639513,0.15037918){\color[rgb]{0,0,0}\makebox(0,0)[lb]{\smash{$5$}}}%
    \put(0.25710028,0.15037918){\color[rgb]{0,0,0}\makebox(0,0)[lb]{\smash{$10$}}}%
    \put(0.34510783,0.15037918){\color[rgb]{0,0,0}\makebox(0,0)[lb]{\smash{$15$}}}%
    \put(0.43107757,0.15037918){\color[rgb]{0,0,0}\makebox(0,0)[lb]{\smash{$20$}}}%
    \put(0.22281518,0.10627529){\color[rgb]{0,0,0}\makebox(0,0)[lb]{\smash{$X$ in [m]}}}%
    \put(0,0){\includegraphics[width=\unitlength,page=5]{OutputExample.pdf}}%
  \end{picture}%
\endgroup%

%% file: Figures/Scenario.pdf_tex
\begingroup%
  \makeatletter%
  \providecommand\color[2][]{%
    \errmessage{(Inkscape) Color is used for the text in Inkscape, but the package 'color.sty' is not loaded}%
    \renewcommand\color[2][]{}%
  }%
  \providecommand\transparent[1]{%
    \errmessage{(Inkscape) Transparency is used (non-zero) for the text in Inkscape, but the package 'transparent.sty' is not loaded}%
    \renewcommand\transparent[1]{}%
  }%
  \providecommand\rotatebox[2]{#2}%
  \ifx\svgwidth\undefined%
    \setlength{\unitlength}{841.88976378bp}%
    \ifx\svgscale\undefined%
      \relax%
    \else%
      \setlength{\unitlength}{\unitlength * \real{\svgscale}}%
    \fi%
  \else%
    \setlength{\unitlength}{\svgwidth}%
  \fi%
  \global\let\svgwidth\undefined%
  \global\let\svgscale\undefined%
  \makeatother%
  \begin{picture}(1,0.70707071)%
    \put(0,0){\includegraphics[width=\unitlength,page=1]{Scenario.pdf}}%
    \put(0.32183468,0.62063808){\color[rgb]{0,0,0}\makebox(0,0)[lb]{\smash{$\text{Vehicle} 1$}}}%
    \put(0.32183468,0.55940482){\color[rgb]{0,0,0}\makebox(0,0)[lb]{\smash{$\text{Vehicle} 2$}}}%
    \put(0.32183468,0.49833524){\color[rgb]{0,0,0}\makebox(0,0)[lb]{\smash{$\text{Bicycle}$}}}%
    \put(0,0){\includegraphics[width=\unitlength,page=2]{Scenario.pdf}}%
    \put(0.5002651,0.00633482){\color[rgb]{0,0,0}\makebox(0,0)[lb]{\smash{$X$ in [m]}}}%
    \put(0.20078927,0.05780385){\color[rgb]{0,0,0}\makebox(0,0)[lb]{\smash{$0$}}}%
    \put(0.28368105,0.05780385){\color[rgb]{0,0,0}\makebox(0,0)[lb]{\smash{$5$}}}%
    \put(0.3590098,0.05780385){\color[rgb]{0,0,0}\makebox(0,0)[lb]{\smash{$10$}}}%
    \put(0.44190153,0.05780385){\color[rgb]{0,0,0}\makebox(0,0)[lb]{\smash{$15$}}}%
    \put(0.52859422,0.05780385){\color[rgb]{0,0,0}\makebox(0,0)[lb]{\smash{$20$}}}%
    \put(0.61148601,0.05780385){\color[rgb]{0,0,0}\makebox(0,0)[lb]{\smash{$25$}}}%
    \put(0.69437773,0.05780385){\color[rgb]{0,0,0}\makebox(0,0)[lb]{\smash{$30$}}}%
    \put(0.77726951,0.05780385){\color[rgb]{0,0,0}\makebox(0,0)[lb]{\smash{$35$}}}%
    \put(0.86016123,0.05780385){\color[rgb]{0,0,0}\makebox(0,0)[lb]{\smash{$40$}}}%
    \put(0.12468637,0.3218336){\color[rgb]{0,0,0}\rotatebox{91.29905303}{\makebox(0,0)[lb]{\smash{$Y$ in [m]}}}}%
    \put(0.16179939,0.09099312){\color[rgb]{0,0,0}\makebox(0,0)[lb]{\smash{$0$}}}%
    \put(0.16179939,0.16227562){\color[rgb]{0,0,0}\makebox(0,0)[lb]{\smash{$5$}}}%
    \put(0.15043544,0.23355813){\color[rgb]{0,0,0}\makebox(0,0)[lb]{\smash{$10$}}}%
    \put(0.15043544,0.30484075){\color[rgb]{0,0,0}\makebox(0,0)[lb]{\smash{$15$}}}%
    \put(0.15043544,0.37612337){\color[rgb]{0,0,0}\makebox(0,0)[lb]{\smash{$20$}}}%
    \put(0.15043544,0.45120685){\color[rgb]{0,0,0}\makebox(0,0)[lb]{\smash{$25$}}}%
    \put(0.15043544,0.52248947){\color[rgb]{0,0,0}\makebox(0,0)[lb]{\smash{$30$}}}%
    \put(0.15043544,0.59377198){\color[rgb]{0,0,0}\makebox(0,0)[lb]{\smash{$35$}}}%
    \put(0.15043544,0.66125362){\color[rgb]{0,0,0}\makebox(0,0)[lb]{\smash{$40$}}}%
  \end{picture}%
\endgroup%

%% file: Figures/histogram2.pdf_tex
\begingroup%
  \makeatletter%
  \providecommand\color[2][]{%
    \errmessage{(Inkscape) Color is used for the text in Inkscape, but the package 'color.sty' is not loaded}%
    \renewcommand\color[2][]{}%
  }%
  \providecommand\transparent[1]{%
    \errmessage{(Inkscape) Transparency is used (non-zero) for the text in Inkscape, but the package 'transparent.sty' is not loaded}%
    \renewcommand\transparent[1]{}%
  }%
  \providecommand\rotatebox[2]{#2}%
  \ifx\svgwidth\undefined%
    \setlength{\unitlength}{820.22870151bp}%
    \ifx\svgscale\undefined%
      \relax%
    \else%
      \setlength{\unitlength}{\unitlength * \real{\svgscale}}%
    \fi%
  \else%
    \setlength{\unitlength}{\svgwidth}%
  \fi%
  \global\let\svgwidth\undefined%
  \global\let\svgscale\undefined%
  \makeatother%
  \begin{picture}(1,0.39731259)%
    \put(0,0){\includegraphics[width=\unitlength]{histogram2.pdf}}%
    \put(0.04914487,0.09441876){\color[rgb]{0.14901961,0.14901961,0.14901961}\makebox(0,0)[lb]{\smash{$0$}}}%
    \put(0.16283213,0.09441876){\color[rgb]{0.14901961,0.14901961,0.14901961}\makebox(0,0)[lb]{\smash{$0.5$}}}%
    \put(0.31396115,0.09441876){\color[rgb]{0.14901961,0.14901961,0.14901961}\makebox(0,0)[lb]{\smash{$1$}}}%
    \put(0.03029767,0.12095781){\color[rgb]{0.14901961,0.14901961,0.14901961}\makebox(0,0)[lb]{\smash{$0$}}}%
    \put(0.00027965,0.20451328){\color[rgb]{0.14901961,0.14901961,0.14901961}\makebox(0,0)[lb]{\smash{$0.6$}}}%
    \put(-0.00026969,0.29197012){\color[rgb]{0.14901961,0.14901961,0.14901961}\makebox(0,0)[lb]{\smash{$1.2$}}}%
    \put(-0.00026969,0.37942694){\color[rgb]{0.14901961,0.14901961,0.14901961}\makebox(0,0)[lb]{\smash{$1.8$}}}%
    \put(0.3859143,0.09481846){\color[rgb]{0.14901961,0.14901961,0.14901961}\makebox(0,0)[lb]{\smash{$0$}}}%
    \put(0.49929567,0.09481846){\color[rgb]{0.14901961,0.14901961,0.14901961}\makebox(0,0)[lb]{\smash{$0.5$}}}%
    \put(0.65030282,0.09481846){\color[rgb]{0.14901961,0.14901961,0.14901961}\makebox(0,0)[lb]{\smash{$1$}}}%
    \put(0.36691376,0.12131514){\color[rgb]{0.14901961,0.14901961,0.14901961}\makebox(0,0)[lb]{\smash{$0$}}}%
    \put(0.33652774,0.20479546){\color[rgb]{0.14901961,0.14901961,0.14901961}\makebox(0,0)[lb]{\smash{$0.6$}}}%
    \put(0.33584696,0.29217713){\color[rgb]{0.14901961,0.14901961,0.14901961}\makebox(0,0)[lb]{\smash{$1.2$}}}%
    \put(0.33584696,0.37955881){\color[rgb]{0.14901961,0.14901961,0.14901961}\makebox(0,0)[lb]{\smash{$1.8$}}}%
    \put(0.72291888,0.09467078){\color[rgb]{0.14901961,0.14901961,0.14901961}\makebox(0,0)[lb]{\smash{$0$}}}%
    \put(0.83619737,0.09467078){\color[rgb]{0.14901961,0.14901961,0.14901961}\makebox(0,0)[lb]{\smash{$0.5$}}}%
    \put(0.98718894,0.09467078){\color[rgb]{0.14901961,0.14901961,0.14901961}\makebox(0,0)[lb]{\smash{$1$}}}%
    \put(0.7038456,0.12131028){\color[rgb]{0.14901961,0.14901961,0.14901961}\makebox(0,0)[lb]{\smash{$0$}}}%
    \put(0.7038456,0.20481539){\color[rgb]{0.14901961,0.14901961,0.14901961}\makebox(0,0)[lb]{\smash{$3$}}}%
    \put(0.7038456,0.29222185){\color[rgb]{0.14901961,0.14901961,0.14901961}\makebox(0,0)[lb]{\smash{$6$}}}%
    \put(0.7038456,0.37962831){\color[rgb]{0.14901961,0.14901961,0.14901961}\makebox(0,0)[lb]{\smash{$9$}}}%
    \put(0.72394065,0.38918538){\color[rgb]{0.14901961,0.14901961,0.14901961}\makebox(0,0)[lb]{\smash{$\times10^4$}}}%
    \put(0.38701593,0.38908635){\color[rgb]{0.14901961,0.14901961,0.14901961}\makebox(0,0)[lb]{\smash{$\times10^4$}}}%
    \put(0.05163909,0.39047969){\color[rgb]{0.14901961,0.14901961,0.14901961}\makebox(0,0)[lb]{\smash{$\times10^4$}}}%
    \put(0.12439082,0.05087484){\color[rgb]{0,0,0}\makebox(0,0)[lb]{\smash{$\text{p}_{q}(o_{t_\text{pred}}^{ij})$ }}}%
    \put(0.0980567,0.00252597){\color[rgb]{0,0,0}\makebox(0,0)[lb]{\smash{for $t_\text{pred}=0.5$}}}%
    \put(0.45405496,0.05087484){\color[rgb]{0,0,0}\makebox(0,0)[lb]{\smash{$\text{p}_{q}(o_{t_\text{pred}}^{ij})$ }}}%
    \put(0.43357287,0.00252597){\color[rgb]{0,0,0}\makebox(0,0)[lb]{\smash{for $t_\text{pred}=1.0$}}}%
    \put(0.79347249,0.05087484){\color[rgb]{0,0,0}\makebox(0,0)[lb]{\smash{$\text{p}_{q}(o_{t_\text{pred}}^{ij})$ }}}%
    \put(0.76908904,0.00252597){\color[rgb]{0,0,0}\makebox(0,0)[lb]{\smash{for $t_\text{pred}=2.0$}}}%
  \end{picture}%
\endgroup%

%% file: Figures/histogram_05.pdf_tex
\begingroup%
  \makeatletter%
  \providecommand\color[2][]{%
    \errmessage{(Inkscape) Color is used for the text in Inkscape, but the package 'color.sty' is not loaded}%
    \renewcommand\color[2][]{}%
  }%
  \providecommand\transparent[1]{%
    \errmessage{(Inkscape) Transparency is used (non-zero) for the text in Inkscape, but the package 'transparent.sty' is not loaded}%
    \renewcommand\transparent[1]{}%
  }%
  \providecommand\rotatebox[2]{#2}%
  \ifx\svgwidth\undefined%
    \setlength{\unitlength}{841.88976378bp}%
    \ifx\svgscale\undefined%
      \relax%
    \else%
      \setlength{\unitlength}{\unitlength * \real{\svgscale}}%
    \fi%
  \else%
    \setlength{\unitlength}{\svgwidth}%
  \fi%
  \global\let\svgwidth\undefined%
  \global\let\svgscale\undefined%
  \makeatother%
  \begin{picture}(1,0.70707071)%
    \put(0.53213618,0.22330717){\color[rgb]{0,0,0}\makebox(0,0)[lb]{\smash{}}}%
    \put(0.12189959,0.18769944){\color[rgb]{0,0,0}\makebox(0,0)[lb]{\smash{${\epsilon}_{{{t_\text{pred}},\text{low}}}$}}}%
    \put(0.45922241,0.18769944){\color[rgb]{0,0,0}\makebox(0,0)[lb]{\smash{${\epsilon}_{{{t_\text{pred}},\text{med}}}$}}}%
    \put(0.77709302,0.18776324){\color[rgb]{0,0,0}\makebox(0,0)[lb]{\smash{${\epsilon}_{{{t_\text{pred}},\text{high}}}$}}}%
    \put(0,0){\includegraphics[width=\unitlength,page=1]{histogram_05.pdf}}%
    \put(0.05834235,0.2372504){\color[rgb]{0.14901961,0.14901961,0.14901961}\makebox(0,0)[lb]{\smash{$0$}}}%
    \put(0.15943734,0.2372504){\color[rgb]{0.14901961,0.14901961,0.14901961}\makebox(0,0)[lb]{\smash{$0.2$}}}%
    \put(0.28103461,0.2372504){\color[rgb]{0.14901961,0.14901961,0.14901961}\makebox(0,0)[lb]{\smash{$0.4$}}}%
    \put(0,0){\includegraphics[width=\unitlength,page=2]{histogram_05.pdf}}%
    \put(0.03557476,0.26970428){\color[rgb]{0.14901961,0.14901961,0.14901961}\makebox(0,0)[lb]{\smash{$0$}}}%
    \put(0.0160021,0.37272377){\color[rgb]{0.14901961,0.14901961,0.14901961}\makebox(0,0)[lb]{\smash{$30$}}}%
    \put(0.0160021,0.48904663){\color[rgb]{0.14901961,0.14901961,0.14901961}\makebox(0,0)[lb]{\smash{$60$}}}%
    \put(0,0){\includegraphics[width=\unitlength,page=3]{histogram_05.pdf}}%
    \put(0.38811827,0.23769251){\color[rgb]{0.14901961,0.14901961,0.14901961}\makebox(0,0)[lb]{\smash{$0$}}}%
    \put(0.48950242,0.23769251){\color[rgb]{0.14901961,0.14901961,0.14901961}\makebox(0,0)[lb]{\smash{$0.2$}}}%
    \put(0.61112675,0.23769251){\color[rgb]{0.14901961,0.14901961,0.14901961}\makebox(0,0)[lb]{\smash{$0.4$}}}%
    \put(0,0){\includegraphics[width=\unitlength,page=4]{histogram_05.pdf}}%
    \put(0.36578755,0.27009126){\color[rgb]{0.14901961,0.14901961,0.14901961}\makebox(0,0)[lb]{\smash{$0$}}}%
    \put(0.34655191,0.37298902){\color[rgb]{0.14901961,0.14901961,0.14901961}\makebox(0,0)[lb]{\smash{$40$}}}%
    \put(0.34655191,0.48919018){\color[rgb]{0.14901961,0.14901961,0.14901961}\makebox(0,0)[lb]{\smash{$80$}}}%
    \put(0,0){\includegraphics[width=\unitlength,page=5]{histogram_05.pdf}}%
    \put(0.70711756,0.23725041){\color[rgb]{0.14901961,0.14901961,0.14901961}\makebox(0,0)[lb]{\smash{$0.05$}}}%
    \put(0.81359574,0.23725041){\color[rgb]{0.14901961,0.14901961,0.14901961}\makebox(0,0)[lb]{\smash{$0.3$}}}%
    \put(0.93297481,0.23725041){\color[rgb]{0.14901961,0.14901961,0.14901961}\makebox(0,0)[lb]{\smash{$0.55$}}}%
    \put(0,0){\includegraphics[width=\unitlength,page=6]{histogram_05.pdf}}%
    \put(0.6901533,0.26970429){\color[rgb]{0.14901961,0.14901961,0.14901961}\makebox(0,0)[lb]{\smash{$0$}}}%
    \put(0.67116627,0.37272378){\color[rgb]{0.14901961,0.14901961,0.14901961}\makebox(0,0)[lb]{\smash{$40$}}}%
    \put(0.67116627,0.48904665){\color[rgb]{0.14901961,0.14901961,0.14901961}\makebox(0,0)[lb]{\smash{$80$}}}%
    \put(0,0){\includegraphics[width=\unitlength,page=7]{histogram_05.pdf}}%
  \end{picture}%
\endgroup%

%% file: Figures/histogram.pdf_tex
\begingroup%
  \makeatletter%
  \providecommand\color[2][]{%
    \errmessage{(Inkscape) Color is used for the text in Inkscape, but the package 'color.sty' is not loaded}%
    \renewcommand\color[2][]{}%
  }%
  \providecommand\transparent[1]{%
    \errmessage{(Inkscape) Transparency is used (non-zero) for the text in Inkscape, but the package 'transparent.sty' is not loaded}%
    \renewcommand\transparent[1]{}%
  }%
  \providecommand\rotatebox[2]{#2}%
  \ifx\svgwidth\undefined%
    \setlength{\unitlength}{841.88976378bp}%
    \ifx\svgscale\undefined%
      \relax%
    \else%
      \setlength{\unitlength}{\unitlength * \real{\svgscale}}%
    \fi%
  \else%
    \setlength{\unitlength}{\svgwidth}%
  \fi%
  \global\let\svgwidth\undefined%
  \global\let\svgscale\undefined%
  \makeatother%
  \begin{picture}(1,0.70707071)%
    \put(0,0){\includegraphics[width=\unitlength,page=1]{histogram.pdf}}%
    \put(0.06025584,0.21904728){\color[rgb]{0.14901961,0.14901961,0.14901961}\makebox(0,0)[lb]{\smash{$0$}}}%
    \put(0.15858285,0.21904728){\color[rgb]{0.14901961,0.14901961,0.14901961}\makebox(0,0)[lb]{\smash{$0.15$}}}%
    \put(0.28562572,0.21904728){\color[rgb]{0.14901961,0.14901961,0.14901961}\makebox(0,0)[lb]{\smash{$0.3$}}}%
    \put(0,0){\includegraphics[width=\unitlength,page=2]{histogram.pdf}}%
    \put(0.04289653,0.25040656){\color[rgb]{0.14901961,0.14901961,0.14901961}\makebox(0,0)[lb]{\smash{$0$}}}%
    \put(0.02456378,0.31690029){\color[rgb]{0.14901961,0.14901961,0.14901961}\makebox(0,0)[lb]{\smash{$20$}}}%
    \put(0.02456378,0.38339398){\color[rgb]{0.14901961,0.14901961,0.14901961}\makebox(0,0)[lb]{\smash{$40$}}}%
    \put(0.02456378,0.44988769){\color[rgb]{0.14901961,0.14901961,0.14901961}\makebox(0,0)[lb]{\smash{$60$}}}%
    \put(0,0){\includegraphics[width=\unitlength,page=3]{histogram.pdf}}%
    \put(0.38563119,0.21555649){\color[rgb]{0.14901961,0.14901961,0.14901961}\makebox(0,0)[lb]{\smash{$0.15$}}}%
    \put(0.49015114,0.21555649){\color[rgb]{0.14901961,0.14901961,0.14901961}\makebox(0,0)[lb]{\smash{$0.35$}}}%
    \put(0.60607401,0.21555649){\color[rgb]{0.14901961,0.14901961,0.14901961}\makebox(0,0)[lb]{\smash{$0.55$}}}%
    \put(0,0){\includegraphics[width=\unitlength,page=4]{histogram.pdf}}%
    \put(0.37247711,0.24985573){\color[rgb]{0.14901961,0.14901961,0.14901961}\makebox(0,0)[lb]{\smash{$0$}}}%
    \put(0.35324147,0.3162564){\color[rgb]{0.14901961,0.14901961,0.14901961}\makebox(0,0)[lb]{\smash{$20$}}}%
    \put(0.35324147,0.38265706){\color[rgb]{0.14901961,0.14901961,0.14901961}\makebox(0,0)[lb]{\smash{$40$}}}%
    \put(0.35324147,0.44905773){\color[rgb]{0.14901961,0.14901961,0.14901961}\makebox(0,0)[lb]{\smash{$60$}}}%
    \put(0,0){\includegraphics[width=\unitlength,page=5]{histogram.pdf}}%
    \put(0.72014677,0.21904728){\color[rgb]{0.14901961,0.14901961,0.14901961}\makebox(0,0)[lb]{\smash{$0.1$}}}%
    \put(0.82454606,0.21904728){\color[rgb]{0.14901961,0.14901961,0.14901961}\makebox(0,0)[lb]{\smash{$0.4$}}}%
    \put(0.9479502,0.21904728){\color[rgb]{0.14901961,0.14901961,0.14901961}\makebox(0,0)[lb]{\smash{$0.7$}}}%
    \put(0,0){\includegraphics[width=\unitlength,page=6]{histogram.pdf}}%
    \put(0.705221,0.25040656){\color[rgb]{0.14901961,0.14901961,0.14901961}\makebox(0,0)[lb]{\smash{$0$}}}%
    \put(0.68688826,0.31690029){\color[rgb]{0.14901961,0.14901961,0.14901961}\makebox(0,0)[lb]{\smash{$20$}}}%
    \put(0.68688826,0.38339398){\color[rgb]{0.14901961,0.14901961,0.14901961}\makebox(0,0)[lb]{\smash{$40$}}}%
    \put(0.68688826,0.44988769){\color[rgb]{0.14901961,0.14901961,0.14901961}\makebox(0,0)[lb]{\smash{$60$}}}%
    \put(0,0){\includegraphics[width=\unitlength,page=7]{histogram.pdf}}%
    \put(0.53213618,0.22330717){\color[rgb]{0,0,0}\makebox(0,0)[lb]{\smash{}}}%
    \put(0.12189959,0.17629652){\color[rgb]{0,0,0}\makebox(0,0)[lb]{\smash{${\epsilon}_{{{t_\text{pred}},\text{low}}}$}}}%
    \put(0.45922241,0.17629652){\color[rgb]{0,0,0}\makebox(0,0)[lb]{\smash{${\epsilon}_{{{t_\text{pred}},\text{med}}}$}}}%
    \put(0.77709302,0.17636032){\color[rgb]{0,0,0}\makebox(0,0)[lb]{\smash{${\epsilon}_{{{t_\text{pred}},\text{high}}}$}}}%
  \end{picture}%
\endgroup%

%% file: Figures/histogram_20.pdf_tex
\begingroup%
  \makeatletter%
  \providecommand\color[2][]{%
    \errmessage{(Inkscape) Color is used for the text in Inkscape, but the package 'color.sty' is not loaded}%
    \renewcommand\color[2][]{}%
  }%
  \providecommand\transparent[1]{%
    \errmessage{(Inkscape) Transparency is used (non-zero) for the text in Inkscape, but the package 'transparent.sty' is not loaded}%
    \renewcommand\transparent[1]{}%
  }%
  \providecommand\rotatebox[2]{#2}%
  \ifx\svgwidth\undefined%
    \setlength{\unitlength}{841.88976378bp}%
    \ifx\svgscale\undefined%
      \relax%
    \else%
      \setlength{\unitlength}{\unitlength * \real{\svgscale}}%
    \fi%
  \else%
    \setlength{\unitlength}{\svgwidth}%
  \fi%
  \global\let\svgwidth\undefined%
  \global\let\svgscale\undefined%
  \makeatother%
  \begin{picture}(1,0.70707071)%
    \put(0.53213618,0.22330717){\color[rgb]{0,0,0}\makebox(0,0)[lb]{\smash{}}}%
    \put(0.12189959,0.19720187){\color[rgb]{0,0,0}\makebox(0,0)[lb]{\smash{${\epsilon}_{{{t_\text{pred}},\text{low}}}$}}}%
    \put(0.45922241,0.19720187){\color[rgb]{0,0,0}\makebox(0,0)[lb]{\smash{${\epsilon}_{{{t_\text{pred}},\text{med}}}$}}}%
    \put(0.77709302,0.19726567){\color[rgb]{0,0,0}\makebox(0,0)[lb]{\smash{${\epsilon}_{{{t_\text{pred}},\text{high}}}$}}}%
    \put(0,0){\includegraphics[width=\unitlength,page=1]{histogram_20.pdf}}%
    \put(0.05400006,0.23476866){\color[rgb]{0.14901961,0.14901961,0.14901961}\makebox(0,0)[lb]{\smash{$0.05$}}}%
    \put(0.15658245,0.23476866){\color[rgb]{0.14901961,0.14901961,0.14901961}\makebox(0,0)[lb]{\smash{$0.09$}}}%
    \put(0.2781697,0.23476866){\color[rgb]{0.14901961,0.14901961,0.14901961}\makebox(0,0)[lb]{\smash{$0.13$}}}%
    \put(0,0){\includegraphics[width=\unitlength,page=2]{histogram_20.pdf}}%
    \put(0.04086311,0.26580737){\color[rgb]{0.14901961,0.14901961,0.14901961}\makebox(0,0)[lb]{\smash{$0$}}}%
    \put(0.01926475,0.3686554){\color[rgb]{0.14901961,0.14901961,0.14901961}\makebox(0,0)[lb]{\smash{$40$}}}%
    \put(0.01926475,0.48480686){\color[rgb]{0.14901961,0.14901961,0.14901961}\makebox(0,0)[lb]{\smash{$80$}}}%
    \put(0,0){\includegraphics[width=\unitlength,page=3]{histogram_20.pdf}}%
    \put(0.38658612,0.23476866){\color[rgb]{0.14901961,0.14901961,0.14901961}\makebox(0,0)[lb]{\smash{$0.1$}}}%
    \put(0.48916851,0.23476866){\color[rgb]{0.14901961,0.14901961,0.14901961}\makebox(0,0)[lb]{\smash{$0.4$}}}%
    \put(0.61075576,0.23476866){\color[rgb]{0.14901961,0.14901961,0.14901961}\makebox(0,0)[lb]{\smash{$0.7$}}}%
    \put(0,0){\includegraphics[width=\unitlength,page=4]{histogram_20.pdf}}%
    \put(0.3709525,0.26580737){\color[rgb]{0.14901961,0.14901961,0.14901961}\makebox(0,0)[lb]{\smash{$0$}}}%
    \put(0.34935414,0.3686554){\color[rgb]{0.14901961,0.14901961,0.14901961}\makebox(0,0)[lb]{\smash{$40$}}}%
    \put(0.34935414,0.48480686){\color[rgb]{0.14901961,0.14901961,0.14901961}\makebox(0,0)[lb]{\smash{$80$}}}%
    \put(0,0){\includegraphics[width=\unitlength,page=5]{histogram_20.pdf}}%
    \put(0.72404883,0.23476866){\color[rgb]{0.14901961,0.14901961,0.14901961}\makebox(0,0)[lb]{\smash{$0$}}}%
    \put(0.8235325,0.23476866){\color[rgb]{0.14901961,0.14901961,0.14901961}\makebox(0,0)[lb]{\smash{$0.5$}}}%
    \put(0.96349653,0.23476866){\color[rgb]{0.14901961,0.14901961,0.14901961}\makebox(0,0)[lb]{\smash{$1$}}}%
    \put(0,0){\includegraphics[width=\unitlength,page=6]{histogram_20.pdf}}%
    \put(0.70551909,0.26580737){\color[rgb]{0.14901961,0.14901961,0.14901961}\makebox(0,0)[lb]{\smash{$0$}}}%
    \put(0.66659985,0.3686554){\color[rgb]{0.14901961,0.14901961,0.14901961}\makebox(0,0)[lb]{\smash{$150$}}}%
    \put(0.66659985,0.48480686){\color[rgb]{0.14901961,0.14901961,0.14901961}\makebox(0,0)[lb]{\smash{$300$}}}%
    \put(0,0){\includegraphics[width=\unitlength,page=7]{histogram_20.pdf}}%
  \end{picture}%
\endgroup%

%% file: Figures/OutputExample2.pdf_tex
\begingroup%
  \makeatletter%
  \providecommand\color[2][]{%
    \errmessage{(Inkscape) Color is used for the text in Inkscape, but the package 'color.sty' is not loaded}%
    \renewcommand\color[2][]{}%
  }%
  \providecommand\transparent[1]{%
    \errmessage{(Inkscape) Transparency is used (non-zero) for the text in Inkscape, but the package 'transparent.sty' is not loaded}%
    \renewcommand\transparent[1]{}%
  }%
  \providecommand\rotatebox[2]{#2}%
  \ifx\svgwidth\undefined%
    \setlength{\unitlength}{841.88976378bp}%
    \ifx\svgscale\undefined%
      \relax%
    \else%
      \setlength{\unitlength}{\unitlength * \real{\svgscale}}%
    \fi%
  \else%
    \setlength{\unitlength}{\svgwidth}%
  \fi%
  \global\let\svgwidth\undefined%
  \global\let\svgscale\undefined%
  \makeatother%
  \begin{picture}(1,0.70707071)%
    \put(0.20830635,0.56814905){\color[rgb]{0,0,0}\makebox(0,0)[lb]{\smash{$\text{Ground Truth}$}}}%
    \put(0.6113569,0.56835338){\color[rgb]{0,0,0}\makebox(0,0)[lb]{\smash{$\text{Output of Machine Learning}$}}}%
    \put(0,0){\includegraphics[width=\unitlength,page=1]{OutputExample2.pdf}}%
    \put(0.59041453,0.12405639){\color[rgb]{0,0,0}\makebox(0,0)[lb]{\smash{$0$}}}%
    \put(0.55924087,0.15802895){\color[rgb]{0,0,0}\makebox(0,0)[lb]{\smash{$0$}}}%
    \put(0.54850025,0.25331562){\color[rgb]{0,0,0}\makebox(0,0)[lb]{\smash{$10$}}}%
    \put(0.548169,0.34782509){\color[rgb]{0,0,0}\makebox(0,0)[lb]{\smash{$20$}}}%
    \put(0.54849061,0.44214462){\color[rgb]{0,0,0}\makebox(0,0)[lb]{\smash{$30$}}}%
    \put(0.54862213,0.53523283){\color[rgb]{0,0,0}\makebox(0,0)[lb]{\smash{$40$}}}%
    \put(0.52970096,0.30033634){\color[rgb]{0,0,0}\rotatebox{89.80085145}{\makebox(0,0)[lb]{\smash{$Y$ in [m]}}}}%
    \put(0,0){\includegraphics[width=\unitlength,page=2]{OutputExample2.pdf}}%
    \put(0.67813818,0.12405639){\color[rgb]{0,0,0}\makebox(0,0)[lb]{\smash{$10$}}}%
    \put(0.7735794,0.12405639){\color[rgb]{0,0,0}\makebox(0,0)[lb]{\smash{$20$}}}%
    \put(0.86902062,0.12405639){\color[rgb]{0,0,0}\makebox(0,0)[lb]{\smash{$30$}}}%
    \put(0.96056737,0.12405639){\color[rgb]{0,0,0}\makebox(0,0)[lb]{\smash{$40$}}}%
    \put(0.75011162,0.08579007){\color[rgb]{0,0,0}\makebox(0,0)[lb]{\smash{$X$ in [m]}}}%
    \put(0,0){\includegraphics[width=\unitlength,page=3]{OutputExample2.pdf}}%
    \put(0.10106169,0.12432124){\color[rgb]{0,0,0}\makebox(0,0)[lb]{\smash{$0$}}}%
    \put(0.06988798,0.15829387){\color[rgb]{0,0,0}\makebox(0,0)[lb]{\smash{$0$}}}%
    \put(0.05914737,0.25358054){\color[rgb]{0,0,0}\makebox(0,0)[lb]{\smash{$10$}}}%
    \put(0.05881612,0.34809001){\color[rgb]{0,0,0}\makebox(0,0)[lb]{\smash{$20$}}}%
    \put(0.05913773,0.44240941){\color[rgb]{0,0,0}\makebox(0,0)[lb]{\smash{$30$}}}%
    \put(0.05926925,0.53549751){\color[rgb]{0,0,0}\makebox(0,0)[lb]{\smash{$40$}}}%
    \put(0.04034863,0.30060103){\color[rgb]{0,0,0}\rotatebox{89.80085145}{\makebox(0,0)[lb]{\smash{$Y$ in [m]}}}}%
    \put(0,0){\includegraphics[width=\unitlength,page=4]{OutputExample2.pdf}}%
    \put(0.18878532,0.12432124){\color[rgb]{0,0,0}\makebox(0,0)[lb]{\smash{$10$}}}%
    \put(0.28422633,0.12432124){\color[rgb]{0,0,0}\makebox(0,0)[lb]{\smash{$20$}}}%
    \put(0.37966755,0.12432124){\color[rgb]{0,0,0}\makebox(0,0)[lb]{\smash{$30$}}}%
    \put(0.47121537,0.12432124){\color[rgb]{0,0,0}\makebox(0,0)[lb]{\smash{$40$}}}%
    \put(0.26075857,0.08605493){\color[rgb]{0,0,0}\makebox(0,0)[lb]{\smash{$X$ in [m]}}}%
    \put(0,0){\includegraphics[width=\unitlength,page=5]{OutputExample2.pdf}}%
  \end{picture}%
\endgroup%